\documentclass[12pt, letterpaper]{article}

\usepackage[sectionbib]{chapterbib}
\usepackage{xcolor}
\usepackage{graphicx}
\usepackage{fullpage}
\usepackage{times}
\usepackage[colorlinks,citecolor=blue,urlcolor=blue,bookmarks=false,hypertexnames=true]{hyperref} 
\usepackage{natbib}
\usepackage{multirow}
\usepackage{footnote}
\usepackage{footmisc}
\usepackage{setspace}
\usepackage{amsmath}
\usepackage{amsfonts}
\usepackage{longtable}
\usepackage[center]{caption}
\usepackage{kbordermatrix}
\def\bibfont{\small}%
\def\bibsep{1pt plus 0.1ex}
\let\oldbibliography\thebibliography
\renewcommand{\thebibliography}[1]{%
	\oldbibliography{#1}%
	\setlength{\itemsep}{0.2pt}%
}

\newcommand{\vectory}{\textbf{\emph{y}}}

\newcommand{\matrixC}{\textbf{\emph{C}}}

\newcommand{\vectord}{\textbf{\emph{d}}}

\newcommand{\vectorr}{\textbf{\emph{r}}}

\newcommand{\matrixA}{\textbf{\emph{A}}}

\newcommand{\vectorb}{\textbf{\emph{b}}}

\newcommand{\vectorbeta}{\boldsymbol{\beta}}

\newcommand{\vectorgamma}{\boldsymbol{\gamma}}

\newcommand{\vectordelta}{\boldsymbol{\delta}}

\newcommand{\vectorp}{\textbf{\emph{p}}}

\newcommand{\matrixQ}{\textbf{\emph{Q}}}

\newcommand{\vectormu}{\boldsymbol{\mu}}

\usepackage{float}
\usepackage{ragged2e}
\usepackage{titlesec}
\usepackage{caption}

\renewcommand{\theequation}{\arabic{equation}}

\usepackage{changepage}
\usepackage{bibunits}

\begin{document}

\textbf{\Large \centering Diversity Preference-Aware Link Recommendation for \\}

\textbf{\Large \centering Online Social Networks\\}

\hspace{1cm}
\begin{center}
Accepted by Information Systems Research\\
\vspace{0.6cm}

Kexin Yin$^1$, Xiao Fang$^{1,2,*}$, Bintong Chen$^{1,2}$, Olivia Sheng$^{3}$ \\
\vspace{0.3cm}
$^1$ Institute for Financial Services Analytics, University of Delaware, Newark, DE, USA \\
$^2$ Lerner College of Business and Economics, University of Delaware, Newark, DE, USA \\
$^3$ David Eccles School of Business, The University of Utah, Salt Lake City, UT, USA \\
\vspace{0.3cm}
$*$ Corresponding Author: Xiao Fang, \href{mailto:xfang@udel.edu}{xfang@udel.edu} 
\end{center}

\hspace{2cm}
\begin{adjustwidth}{0.6cm}{0.6cm}

\noindent \textbf{Abstract:} Link recommendation, which recommends links to connect unlinked online social network users, is a fundamental social network analytics problem with ample business implications. Existing link recommendation methods tend to recommend similar friends to a user but overlook the user's diversity preference, although social psychology theories suggest the criticality of diversity preference to link recommendation performance. In recommender systems, a field related to link recommendation, a number of diversification methods have been proposed to improve the diversity of recommended items. Nevertheless, diversity preference is distinct from diversity studied by diversification methods. To address these research gaps, we define and operationalize the concept of diversity preference for link recommendation and propose a new link recommendation problem: the diversity preference-aware link recommendation problem. We then analyze key properties of the new link recommendation problem and develop a novel link recommendation method to solve the problem. Using two large-scale online social network data sets, we conduct extensive empirical evaluations to demonstrate the superior performance of our method over representative diversification methods adapted for link recommendation as well as state-of-the-art link recommendation methods.

\hspace{1cm} 

\end{adjustwidth}
\hspace{1cm}

\section{Introduction}

Owing to the increasing Internet penetration worldwide, the number of social media users and their activities on the media have grown rapidly in recent years. The annual revenue generated worldwide by major social media firms reached \$92.89 billion in 2019,\footnote{Revealed: The social media platforms that make the most revenue off their users. Retrieved on December 29, 2020 from \url{https://www.digitalinformationworld.com/2019/12/revenue-per-social-media-user.html}.} and there were 3.8 billion active social media users at the beginning of 2020, accounting for 49$\%$ of the world's population.\footnote{Digital 2020: Global Digital Overview. Retrieved on December 29, 2020 from  \url{https://datareportal.com/reports/digital-2020-global-digital-overview}.} Online social networks, such as Facebook and LinkedIn, constitute a significant portion of social media. These social networks have become an essential part of many people's daily lives, from socializing with friends on Facebook to seeking job opportunities on LinkedIn. Recognizing the widespread use of online social networks, researchers have been actively developing advanced predictive analytics methods that leverage these networks for greater economic and societal benefits  \citep{kempe2003maximizing,fang2013predicting,li2020will}. In this vein, prior studies have proposed predictive analytics methods to solve important online social network problems with significant business implications, ranging from adoption probability and top persuader predictions \citep{fang2013predicting,fang2016top} to link recommendation and social foci's popularity prediction \citep{li2017utility,li2020will}. In particular, designing predictive analytics methods for link recommendation has attracted attention from both academic researchers and business practitioners \citep{davenport2012data,li2017survey}.

Link recommendation, which recommends links to connect currently unlinked users, is a key functionality commonly offered by online social networks \citep{li2017survey}. Notable examples of link recommendation include friend recommendation features (e.g., ``People you may know'') on Facebook and LinkedIn. Effective link recommendation caters to users' needs by enlarging their social circles in online social networks, thereby improving their experience and satisfaction with those networks \citep{li2017utility}. Moreover, effective link recommendation leads to a more densely connected online social network, which better facilitates information and advertisement diffusion in the network and creates more revenue (e.g., advertisement revenue) for the network operator (e.g., Facebook Inc.) \citep{shriver2013social,li2017utility}. Existing link recommendation
methods predict the linkage likelihood between a pair of currently unlinked users based on their degree of similarity---for example, similarity in their profiles---and recommend
links with the highest linkage likelihoods \citep{liben2007link,backstrom2011supervised,grover2016node2vec,pareja2020evolvegcn}. Consequently, friends recommended by these methods tend to be similar to each other in terms of their profiles. 


However, while some users are satisfied with homogeneous friend recommendations, others prefer recommended friends with diverse profiles.
For example, people who feel comfortable with familiar things prefer to befriend those with similar profiles, while people who enjoy new things tend to connect with friends of diverse backgrounds.
That is, different users have different diversity preferences when making friends in a social network \citep{burt1998personality,laakasuo2017company}. Theoretically, friendship formation between two users depends on the compatibility and complementarity of their profiles \citep{rivera2010dynamics}. The former, known as homophily, refers to people's tendency to connect with those similar to them \citep{rivera2010dynamics}. The latter, known as heterophily, stipulates that people benefit from connecting with those who are complementary to them \citep{rivera2010dynamics}. Homophily leads to homogeneity among a user's friends, while heterophily promotes heterogeneity among these friends \citep{laakasuo2017company}. Owing to different personalities, different users value homophily and heterophily differently, thereby exhibiting different diversity preferences \citep{laakasuo2017company}. For example, users with a high level of openness prefer heterophily over homophily and tend to befriend those with diverse backgrounds \citep{laakasuo2017company}, because they appreciate new things and believe that similar friends produce only marginal returns \citep{sibley2008personality,tulin2018personality}. In contrast, users with a low level of openness place a greater emphasis on homophily and prefer to connect with those who have similar backgrounds \citep{laakasuo2017company}, as they are more comfortable with familiar things and thus more likely to have relationships with those who have similar profiles \citep{tulin2018personality}. As another example, \cite{burt1998personality} find that people who seek change and authority prefer heterophily and tend to connect with others who have diverse profiles, while people who enjoy conformity, obedience, security, and stability prefer homophily and are inclined to form friendships with others who have similar profiles.



In short, social psychology theories suggest that different users have different diversity preferences for forming friendships in a social network \citep{burt1998personality, rivera2010dynamics, laakasuo2017company}. Therefore, an effective link recommendation method should take a user's diversity preference into account and recommend friends that satisfy that preference. As a result, the recommended friends will better meet the user's preference and thus are more likely to be accepted by the user. However, existing link recommendation methods overlook diversity preference. In a field related to link recommendation, recommender systems, a number of diversification methods have been proposed to improve the diversity of recommended items \citep{ziegler2005improving,boim2011diversification,chen2018fast}. Nevertheless, diversity preference, as researched in our study, is distinct from diversity studied by diversification methods in the following ways. First, diversification methods indiscriminately increase the diversity of items recommended by a recommender system for every user \citep{ziegler2005improving,boim2011diversification,chen2018fast}, while we aim to develop a link recommendation method that meets each individual user's diversity preference for forming friendships in a social network. That is, an ideal link recommendation method should provide a personalized solution that will increase or decrease the diversity of recommended friends in accordance with each user's diversity preference. Second, social psychology literature notes that a user may seek friends who are similar on some profile dimensions (e.g., college major) but dissimilar on others (e.g., school) \citep{rivera2010dynamics}. For example, a user may prefer befriending those who are studying Information Systems at a diverse set of universities. Accordingly, users' diversity preference should be measured and optimized at the dimension level, in contrast to diversity that is measured and increased at the user level by diversification methods \citep{ziegler2005improving,boim2011diversification,chen2018fast}.

To address the research gaps discussed above, our study contributes to the literature by proposing a new link recommendation problem and a novel link recommendation method, both of which take diversity preference into account. Specifically, we define the concept of diversity preference in the context of link recommendation and propose how to measure it at the dimension level. Building on this concept, we formally define the diversity preference-aware link recommendation problem. Our link recommendation problem differs from existing link recommendation problems in its consideration of diversity preference. It is also different from diversification problems that are studied in the field of recommender systems because of the difference between diversity preference in our problem and diversity in diversification problems. Furthermore, we analyze key properties of the diversity preference-aware link recommendation problem and propose a novel link recommendation method that solves the problem based on those properties. The key methodological novelty of our method lies in its optimization of each individual user's diversity preference at the dimension level. 
Through extensive experiments with two real-world social network data sets, we demonstrate the superiority of our method over representative diversification methods adapted for link recommendation as well as state-of-the-art link recommendation methods, in terms of recommendation accuracy and the extent to which recommended friends satisfy a user's diversity preference.


\section{Related Work}

Two streams of research are closely related to our study: link recommendation for social networks and diversification methods for recommender system. In this section, we review representative studies in each stream and highlight the key novelties of our study. 

\subsection{Link Recommendation for Social Networks}

In general, prior research treats link recommendation as a supervised learning problem, for which training data is constructed from observed link establishments. Each record of training data consists of predictors that affect link establishment between a pair of users and a class label that is set to 1 if a link connecting the user pair exists or 0 otherwise. Predictors are computed from nodal and/or structural proximities between users. 
Specifically, nodal proximity measures the similarity between users based on their profile attributes, such as age, education, and occupation, using similarity functions, such as cosine similarity and Jaccard's coefficient \citep{zheleva2008using, schifanella2010folks, wang2011human}.
Structural proximity also measures the similarity between users but based on their structural features in a social network \citep{liben2007link}. Representative structural proximity measures include common neighbors, Adamic/Adar, and the Katz index \citep{li2017survey}. 
In particular, common neighbors between a pair of users are calculated as the number of their mutual friends in the social network. Adamic/Adar, an extension of common neighbors,  weights mutual friends differently according to their number of connections in the social network. While common neighbors and Adamic/Adar target users' neighborhoods, the Katz index focuses on paths connecting users. Specifically, the Katz index between a pair of users is computed as the number of paths connecting them, each of which is inversely weighted by its length.

Once training data is constructed, machine learning methods can be applied to the data to build a model, which in turn predicts the linkage likelihood for each pair of currently unlinked users. Machine learning methods commonly used for link recommendation include boosted decision tree \citep{benchettara2010supervised}, supervised random walk \citep{backstrom2011supervised}, support vector machine (SVM) \citep{GongSVM}, and Bayesian network \citep{li2017utility}. For example, \cite{GongSVM} apply a SVM to an attribute-augmented social network to predict linkage likelihoods of Google+ users. \cite{backstrom2011supervised} formulate the link recommendation problem as a supervised learning task with the objective of learning the strength of each link such that a random walk starting from a focal user will more likely visit users who directly connect to the focal user than those who do not. A recent study by \cite{li2017utility} proposes a Bayesian network method that considers both the likelihood and utility of linkage. 

Link recommendation methods reviewed above rely on hand-designed user features for link recommendations. Powered by recent advancements in deep learning, current link recommendation methods employ representation learning that learns to represent a user as a low-dimensional numerical vector with the objective of preserving the user's structural and nodal characteristics in a social network. Pairs of these learned vectors can then be fed into a machine learning algorithm (e.g., logistic regression) for linkage likelihood prediction. In this vein, network embedding methods, such as DeepWalk \citep{perozzi2014deepwalk} and Node2vec \citep{grover2016node2vec}, re-purpose SkipGram, a popular word embedding method in natural language processing \citep{mikolov2013efficient}, to represent nodes (users) in a social network. 
As an example, DeepWalk defines each node as a ``word" and a random walk sequence from the node as its ``context." With these definitions, SkipGram can be applied to derive the representation of each node in a social network.
Although popular for node representations, network embedding methods have limitations: (1) they cannot produce representations for nodes that are not seen during training, and (2) they focus on representing structural features of nodes but overlook their intrinsic features (e.g., user profiles) \citep{hamilton2017representation}. State-of-the-art graph neural networks address these limitations by generating node representations through neural networks, e.g., convolutional neural networks \citep{kipf2016semi, velivckovic2017graph, pareja2020evolvegcn, wu2020comprehensive}. The graph convolutional network (GCN) proposed by \cite{kipf2016semi} is a representative of this category. Rooted in spectral graph theory, GCN is an efficient convolutional neural network that can embed both structural and intrinsic features of a node in a social network. 
In sum, existing link recommendation methods rely on similarities between users for friend recommendations. Consequently, these methods tend to recommend friends who are similar to each other and neglect the diversity preference of individual users.

\subsection{Diversification Methods for Recommender Systems}

While diversity preference is overlooked by existing link recommendation methods, diversity has been studied in the related area of recommender systems \citep{castells2015novelty, kaminskas2016diversity, wu2019recent}. Recommender systems recommend items (e.g., books) to users \citep{adomavicius2005toward}, whereas link recommendation methods recommend users (e.g., friends) to users \citep{li2017survey}. Given the shared commonality of recommendations between these two fields, diversification methods developed for recommender systems can be adapted for link recommendation. 

Classical recommender systems aim to maximize only the accuracy of recommendations \citep{adomavicius2005toward}. As a result, items recommended by these systems tend to be similar to each other \citep{zhang2009novel}. Recognizing the importance of the diversity of recommended items, a number of  diversification methods have been proposed for recommender systems \citep{ziegler2005improving,ekstrand2014user}. The majority of these methods re-rank the items generated by an accuracy-maximizing recommender system to meet the diversity requirement \citep[e.g.,][]{ziegler2005improving,zhang2008avoiding,chen2018fast}, and hence they are generic and can be integrated with any recommender system \citep{chen2018fast}. In the following, we focus our review on diversification methods that are based on re-ranking, as these can also be combined with any link recommendation algorithm to produce diversified link recommendations. 

Most re-ranking-based diversification methods balance the trade-off between recommendation accuracy and diversity because improving one usually sacrifices the other \citep{kaminskas2016diversity}. Accordingly, the objective function of these methods is formulated as a weighted sum of recommendation accuracy and diversity, where the weights control the recommender system's preferences for accuracy and diversity. Recommendation diversity is normally measured as the average pair-wise dissimilarity of recommended items \citep{ziegler2005improving, zhang2008avoiding}. Motivated by the maximal marginal relevance method in information retrieval, \cite{ziegler2005improving} propose a greedy diversification method that selects one item at a time for the final recommendation list. Specifically, an item is selected if it maximizes the weighted sum of its relevance score (i.e., recommendation accuracy) and its average dissimilarity to the items already in the final recommendation list (i.e., recommendation diversity). 
Rather than using a greedy heuristic, \cite{zhang2008avoiding} model the recommendation diversification problem as an optimization problem that maximizes the weighted sum of recommendation accuracy and diversity given the constraint of recommending a predetermined number of items. The diversified recommendation list can then be derived by solving the optimization problem. 
\cite{chen2018fast} propose a new diversity measure that is based on relationships among all recommended items rather than their pair-wise relationships. Built on an elegant probabilistic model -- the determinantal point process, they design a greedy algorithm to search for the final recommendation list. Instead of balancing the trade-off between recommendation accuracy and diversity, \cite{boim2011diversification} adopt a different perspective for diversifying recommended items. They propose first clustering items produced by an accuracy-maximizing recommender system based on their pair-wise dissimilarities. Then one item with the highest relevance score is selected from each cluster to form the final recommendation list. 

In addition, various types of preferences have been utilized by preference-aware recommendation methods to improve the performance of recommendations \citep{rendle2009bpr}. In this vein, \cite{debnath2016preference} consider users' location and temporal preferences for point of interest (POI) recommendations; \cite{zangerle2020user} leverage users' cultural preferences for music recommendations; \cite{shi2012adaptive} and \cite{su2013set} diversify items recommended to a user based on the user's preferred diversification level. Among preference-aware recommendation methods, those considering preferences for diversification levels, such as \cite{shi2012adaptive} and \cite{su2013set}, are most relevant to our study. These methods increase the diversity of a recommendation list based on a user's preferred diversification level (i.e., how much diversity should be increased) \citep{shi2012adaptive, su2013set}.


Our literature review reveals several research gaps. First, existing link recommendation methods fail to consider diversity preference. However, social psychology theories suggest that users exhibit different diversity preferences when forming friendships in a social network \citep{burt1998personality, rivera2010dynamics, laakasuo2017company}. Thus, an effective link recommendation method should take diversity preference into account. Second, although diversification methods and preference-aware methods for recommender systems can be adapted for link recommendation, these methods indiscriminately increase the diversity of recommended friends for all users rather than satisfying each individual user's diversity preference. Third, diversification methods and preference-aware methods measure and increase diversity at the user level, while the literature suggests that diversity preference should be measured and optimized at the dimension level \citep{rivera2010dynamics}. To address these research gaps, we introduce the concept of diversity preference, propose how to measure it at the dimension level, and define a new link recommendation problem: the diversity preference-aware link recommendation problem. We then propose a novel link recommendation method to solve the problem.

\section{Problem Formulation}

We first define the concept of diversity preference and then formulate the diversity preference-aware link recommendation problem. Let $\mathcal{G}=<\mathbb{U},\mathbb{E}>$ be an online social network of $n$ users, where $\mathbb{U} = \{u_i\}$, $i=1,2,\cdots,n$, represents the set of users in the network and $\mathbb{E}$ denotes the set of existing links (e.g., friendships) connecting the users. Let $\mathbb{F}^i=\{u_f \  | \ (u_f,u_i )\in \mathbb{E},\ f\neq i,\:f=1,2,\cdots,n\}$ represent $u_i$'s friends. Each user is described by a $H$-dimensional user profile, e.g., $<major, school, employer>$. The profiles of a user's existing friends reflect the user's diversity preference in making friends \citep{tropp2006valuing, zeng2008preference}. For example, if 24 out of Karen's 30 friends major in Information Systems (IS), this indicates
Karen's preference for low diversity in terms of major as well as her strong preference for befriending those studying IS. On the other hand, if John's friends are evenly distributed over many different majors, this signals that John prefers high  diversity in terms of major. A link recommendation method can leverage a user's diversity preference and recommend friends to best match this preference, thereby recommending friends who are more likely to be accepted by the user. Formally, we define a user's diversity preference as follows.

{\raggedright \textbf{Definition 1. Diversity Preference.}} Given an online social network $\mathcal{G}=(\mathbb{U},\mathbb{E})$ and $H$-dimensional profiles of the users in $\mathbb{U}$, user $u_i$'s diversity preference for dimension $h,\ h=1,\ 2,\cdots, H$, is a $Z_h$-dimensional vector $\vectord^{i,h} \in \mathbb{R}^{Z_h}$, where $Z_h$ is the number of possible values of dimension $h$.	The $z$th element of $\vectord^{i,h}$, $d^{i,h}_z$, represents $u_i$'s preference regarding the $z$th value of dimension $h$ and is measured as the number of $u_i$'s friends who take that value in dimension $h$.\footnote{Depending on the application context, a user's diversity preference may be computed based on all her friends or only her recently acquired friends. Our definition of diversity preference as well as our proposed method to solve the diversity preference-aware link recommendation problem are applicable to both scenarios.}

{\raggedright \textbf{Example 1.}}
Consider Karen, who has 30 friends in an online social network. The \emph{major} dimension has 90 different values (i.e., $Z_{major}=90$), such as IS, Computer Science (CS), Math, and Finance (Fin). Among Karen's 30 friends, 24 major in IS, 3 major in CS, and the remaining 3 double major in Math and CS. According to Definition 1, Karen's diversity preference for the \emph{major} dimension $\vectord^{Karen, major}$ is given by

\vspace{-0.2cm}

\begin{figure}[H]
	
	\centering
	\includegraphics[width=10cm]{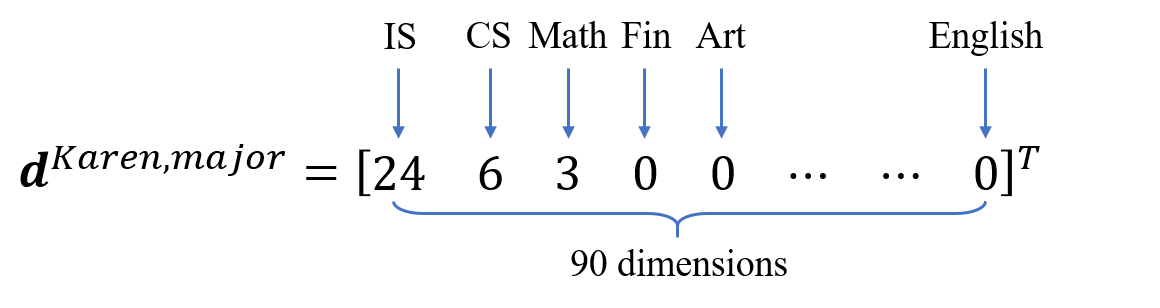}
\end{figure}

\vspace{-0.2cm}

\noindent where the superscript $T$ denotes the transpose of a vector.

User $u_j$ is a potential friend of $u_i$ if she is not a current friend of $u_i$, i.e., $u_j \not \in \mathbb{F}^i$. The linkage likelihood between $u_i$ and $u_j$, $ll(u_i,u_j)$, is the likelihood that $u_i$ and $u_j$ will become friends in the future, which can be estimated using an existing link prediction method \citep{li2017survey}. A candidate friend of $u_i$ is a potential friend of $u_i$ with a relatively high linkage likelihood. More precisely, we define $\mathbb{C}^i=\{u_j \ | \ rank(ll(u_i,\,u_j))\leq m, \ u_j \not \in \mathbb{F}^i,\ j=1, 2, \cdots, n\}$ as the set of $u_i$'s candidate friends, which contains $u_i$'s top-$m$ potential friends as ranked by linkage likelihood. The function $rank(\cdot)$ returns the rank of $u_i$'s likelihood of befriending $u_j$ among the linkage likelihoods between $u_i$ and each of $u_i$'s potential friends. Given candidate set $\mathbb{C}^i$ and the profile of each candidate in $\mathbb{C}^i$, we can construct a candidate profile matrix $\matrixC^{i,h}\in \mathbb{R}^{Z_h \times |\mathbb{C}^i|}$ for each profile dimension $h$, $h=1,2,\cdots,H$, where $Z_h$ is the number of possible values of dimension $h$ and $|\mathbb{C}^i|$ is the number of $u_i$'s candidate friends. An element $\matrixC^{i,h}_{z,q}$ of $\matrixC^{i,h}$ is set to 1 if candidate $q$ takes the $z$th value in dimension $h$ or set to 0 otherwise.

{\raggedright \textbf{Example 2.}}
Consider Karen and her candidate friend set $\mathbb{C}^{Karen}=\{u_1,\ u_2,\ u_3,\ u_4,\ u_5,\ u_6\}$. The \emph{major} dimension has 90 different values, i.e., $Z_{major}=90$. Among Karen's candidate friends, $u_1$ and $u_4$ major in IS, $u_2$ majors in CS, $u_3$ majors in Math, $u_5$ double majors in IS and Finance, and $u_6$ double majors in CS and Math. Karen's candidate profile matrix for the \emph{major} dimension, $\matrixC^{Karen,\ major} \in \mathbb{R}^{90 \times 6}$, is as follows:
\vspace{-5mm}
\[ \matrixC^{Karen,\ major}=\kbordermatrix{
	& u_1 & u_2 & u_3 & u_4 & u_5 & u_6 \\
	IS      & 1 & 0 & 0 & 1 & 1 & 0 \\
	CS      & 0 & 1 & 0 & 0 & 0 & 1 \\
	Math    & 0 & 0 & 1 & 0 & 0 & 1 \\
	Finance & 0 & 0 & 0 & 0 & 1 & 0 \\
	Art     & 0 &   & \cdots & \cdots &  & 0 \\
	\vdots  & \vdots &   & \ddots   &   &   & \vdots \\
	\vdots  & \vdots  &   &   & \ddots  &   & \vdots \\
	English & 0 &   & \cdots & \cdots &  & 0 
}\]
\vspace{-4mm}

Given $u_i$'s candidate set $\mathbb{C}^i$ and the number $k$ of friends to be recommended to $u_i$, the objective
of the diversity preference-aware link recommendation problem is to select $k$ candidates from $\mathbb{C}^i$ that
best satisfy $u_i$'s diversity preference, where $k < |\mathbb{C}^i|$. Concretely, the objective is to maximize the
similarity between $u_i$'s diversity preference and the diversity of the $k$ candidate friends recommended
to $u_i$ across all profile dimensions.
Let $\vectory^i=[y_1^i \ \cdots \ y^i_{|\mathbb{C}^i|}]^T$ be the recommendation decision vector, where $y^i_j=1$ if the $j^{th}$ candidate in $\mathbb{C}^i$ is recommended to $u_i$ or $y^i_j=0$ otherwise, $j=1,2,\cdots, |\mathbb{C}^i|$.
The diversity distribution of candidates recommended to $u_i$ for dimension $h$, $\vectorr^{i,h}\in \mathbb{R}^{Z_h}$, can be calculated as
\begin{equation*}
\vectorr^{i,h} = \matrixC^{i,h}\vectory^i.
\end{equation*}

{\raggedright \textbf{Example 3.}}
Consider Karen's candidate set $\mathbb{C}^{Karen}$ and her candidate profile matrix for the \emph{major} dimension $\matrixC^{Karen,major}$ described in Example 2. Let $\vectory^{Karen}=[1\ 0 \ 0\ 1 \ 1\ 1 ]^T$, i.e., candidates $u_1$, $u_4$, $u_5$, and $u_6$ are recommended to Karen. The diversity distribution of these recommended candidates for the \emph{major} dimension, $\vectorr^{Karen, major}$, is calculated as:

\vspace{-8mm}
\[ \vectorr^{Karen,\ major}=\matrixC^{Karen,\ major}\vectory^{Karen}=
\kbordermatrix{
	& u_1 & u_2 & u_3 & u_4 & u_5 & u_6 \\
	IS      & 1 & 0 & 0 & 1 & 1 & 0 \\
	CS      & 0 & 1 & 0 & 0 & 0 & 1 \\
	Math    & 0 & 0 & 1 & 0 & 0 & 1 \\
	Finance & 0 & 0 & 0 & 0 & 1 & 0 \\
	Art     & 0 &   & \cdots & \cdots &  & 0 \\
	\vdots  & \vdots &   & \ddots   &   &   & \vdots \\
	\vdots  & \vdots  &   &   & \ddots  &   & \vdots \\
	English & 0 &   & \cdots & \cdots &  & 0 
}\kbordermatrix{  
	\\
	u_1   & 1  \\
	u_2   & 0  \\
	u_3   & 0  \\
	u_4   & 1  \\
	u_5   & 1  \\
	u_6   & 1  \\
}=\kbordermatrix{
	\\
	IS      & 3 \\
	CS      & 1 \\
	Math    & 1 \\
	Finance & 1 \\
	Art     & 0 \\
	\vdots  & \vdots \\
	\vdots  & \vdots \\
	English & 0 
}.\]
\vspace{-3mm}

\noindent That is, among the 6 candidate friends recommended to Karen, 3 major in IS, 1 in CS, 1 in Math, and 1 in Finance. 

Given $u_i$'s diversity preference for dimension $h$, $\vectord^{i,h}$, and the diversity distribution of the candidates recommended to $u_i$ for the same dimension, $\vectorr^{i,h}$, a natural similarity function between them is the cosine similarity function:
\[cos(\vectord^{i,h},\vectorr^{i,h}) =  \frac{ \vectord^{i,h^T} \vectorr^{i, h}}{\| \vectord^{i,h} \| \|\vectorr^{i, h}\|},\] 
where $\|\cdot\|$ denotes the L2 norm of a vector.\footnote{In the calculation of the cosine similarity, each vector is normalized by dividing its L2 norm. As a result, cosine similarity cannot differentiate vectors of the same direction but with different magnitudes. To address this limitation of the cosine similarity function, we can drop its normalization step. However, a user has multiple profile dimensions (e.g., major, school, place lived), each of which is described by a diversity preference vector. Therefore, keeping the normalization step is essential to enable the calculated similarities comparable across different dimensions. Otherwise, similarities of some dimensions could substantially outweigh (even dominate) similarities of some other dimensions.} We are now ready to define the diversity preference-aware link recommendation problem.

{\raggedright \textbf{Definition 2. Diversity Preference-Aware Link Recommendation (DPA-LR) Problem.}}
Given a social network $\mathcal{G}=(\mathbb{U},\mathbb{E})$,  the $H$-dimensional profiles of users in $\mathbb{U}$, and a candidate set $\mathbb{C}^i$ for user $u_i$, recommend $k$ friends  from $\mathbb{C}^i$ to $u_i$, $k<|\mathbb{C}^i|$, such that the total similarity between $u_i$'s diversity preference and the diversity distribution of the $k$ recommended friends across all $H$ dimensions, $\sum_{h=1}^{H} \frac{ \vectord^{i,h^T} \vectorr^{i, h}}{\| \vectord^{i,h} \| \|\vectorr^{i, h}\|}$, is maximized.

For the convenience of the reader, important notation used in this paper is listed in Table 1.
\begin{table}[h!]
	\begin{center}
	{\small
		\caption{Notation}
		\label{tab:table1}
		\begin{tabular}{c l} \hline
			
			\multicolumn{1}{c}{\textbf{\small{Notation}}} & \multicolumn{1}{c}{\textbf{\small{Description}}}  \\\hline
			
			$\mathcal{G}$  & An online social network  \\

			$\mathbb{U}$  & The set of users in $\mathcal{G}$   \\

			$\mathbb{E}$  & The set of links in $\mathcal{G}$   \\

			$\mathbb{F}^i$ & The set of user $u_i$'s existing friends    \\

			$\mathbb{C}^i$ & The set of user $u_i$'s candidate friends    \\

			$m$ & The number of a user's candidate friends    \\

			$k$ & The number of candidate friends recommended to a user  \\

			$\vectord^{i,h}$ & User $u_i$'s diversity preference for dimension $h$; see Definition 1    \\

			$\matrixC^{i,h}$ & User $u_i$'s candidate profile matrix for dimension $h$; see Example 2   \\

			$\vectory^i$ & Recommendation decision vector for user $u_i$   \\

			$\vectorr^{i,h}$ & Diversity distribution of candidate friends recommended to $u_i$ for dimension $h$; \\ [-0.1em]
			 & see Example 3
			\\\hline
			
		\end{tabular}}
	\end{center}
\end{table}

\section{Method}

By definition, the DPA-LR problem for a given user can be formulated as the following optimization problem:
\begin{equation}
\begin{aligned}
& \underset{\vectory}{\text{maximize}} 
& & \sum_{h=1}^{H} \frac{\vectord^{h^T} \vectorr^h}{\|\vectord^h\| \|\vectorr^h\|} \\
& \text{subject to} 
& & \textbf{1}^T\vectory=k   \\
&&& y_j \in \{0, 1\},\ j=1,2,\cdots,m
\end{aligned}
\end{equation}

\noindent where $\vectorr^h=\matrixC^h\vectory$, $\matrixC^h$ is the user's candidate profile matrix for dimension $h$, $m$ is the number of the user's candidate friends, $\vectory=[y_1,y_2,\cdots,y_m]^T$ is the decision vector, and $y_j=1$ if the $j^{th}$ candidate is recommended to the user or $y_j=0$ otherwise. 
Notice that we drop the user index $i$ in problem (1) since the optimization formulation is the same for every user.
Problem (1) is a nonlinear sum-of-ratios binary integer programming problem because its objective function is a summation of non-linear fractions and the domain of its decision variables is $\{0,1\}$ \citep{schaible2003fractional, chen2010applied}.
It belongs to a class of problems that are difficult to solve, as demonstrated by the following result on its computational complexity. 

\textbf{Theorem 1.} The diversity preference-aware link recommendation (DPA-LR) problem is NP-hard.

\textbf{Proof.} See Appendix A.

A common approach to solving many NP-hard problems is to relax the original problem and try to obtain an approximate solution. We proceed by relaxing the binary integer constraints $y_j\in\{0,1\}$ in problem (1) to $0 \leq y_j\leq1$, a common approximation practice for NP-hard integer programming problems \citep{karlof2005integer}. 
In addition, we simplify the notation for the problem by defining $\bar{\vectord}^h = \frac{\vectord^h}{\|\vectord^h\|}$ and write out $\vectorr^h$ as $\matrixC^h\vectory$. 
We then obtain the following nonlinear continuous optimization problem as a relaxation of problem (1):
\begin{equation}
\begin{aligned}
& \underset{\vectory}{\text{maximize}} 
& & \sum_{h=1}^{H} \frac{\bar{\vectord}^{h^T} \matrixC^h\vectory}{\|\matrixC^h\vectory \|} \\
& \text{subject to} 
& & \textbf{1}^T\vectory-k=0 \\
&&& \matrixA\vectory-\vectorb \preceq \textbf{0}
\end{aligned}
\end{equation}

\vspace{0.2cm}

\noindent where $\matrixA = [I, -I]^T$, $I\in\mathbb{R}^{m\times m}$ is an identity matrix, $\vectorb=[\boldsymbol{1}^T, \boldsymbol{0}^T]^T$, $\boldsymbol{1}\in \mathbb{R}^m$, $\boldsymbol{0}\in\mathbb{R}^m$, and the notation $\preceq$ denotes vector inequality. Note that the only difference between problem (1) and problem (2) is the relaxation of the constraints on the decision variables, i.e., $y_j\in\{0,1\}$ in problem (1) is relaxed to $0 \leq y_j\leq1$ in problem (2), $j=1,2,\cdots , m$.  

The objective function of problem (2) is non-concave since it is a summation of $H$ non-concave fractional functions $\frac{\bar{\vectord}^{h^T} \matrixC^h\vectory} {\|\matrixC^h\vectory \|}$, $h=1,2,\cdots, H$. As a result, problem (2) is a non-convex optimization problem\footnote{A convex optimization problem minimizes (maximizes) a convex (concave) objective function on a convex set \citep{boyd2004convex}.} and is therefore still difficult to solve \citep{benson2002global}. To tackle the problem, we introduce a variable substitution: $\vectorbeta =[\beta_1,\beta_2,\cdots,\beta_H]^T$, $\beta_h\geq 0 \ \text{for} \ h=1,2,\cdots,H$, such that
\begin{equation*}
\beta_h = \frac{\bar{\vectord}^{h^T} \matrixC^h\vectory} {\|\matrixC^h\vectory \|} \hspace{5mm} \text{for} \hspace{2mm} h=1,2,\cdots, H.
\end{equation*}

\noindent Using the new variable in the objective function and adding the variable substitution in the constraint, we obtain the following reformulation that is equivalent to problem (2): 
\begin{equation}
\begin{aligned}
& \underset{\vectory,\vectorbeta}{\text{maximize}}
& & \textbf{1}^T\vectorbeta \\
& \text{subject to}
& & \textbf{1}^T\vectory-k=0 \\
&&& \matrixA\vectory-\vectorb \preceq \textbf{0} \\
&&& \beta_h\|\matrixC^h\vectory\|-\bar{\vectord}^{h^T}\matrixC^h\vectory = 0, \ h=1,2,\cdots, H
\end{aligned}
\end{equation}

Note that problem (3) remains a non-convex optimization problem, because $\vectorbeta$ is a new decision variable and its last set of constraints are non-convex with respect to $\vectory$ and $\vectorbeta$. To resolve the non-convexity, we introduce Lagrangian multipliers $\vectorgamma =[\gamma_1,\gamma_2,\cdots,\gamma_H]^T$ for the last set of constraints in problem (3) and focus on the following Lagrangian optimization problem:
\begin{equation}
\begin{aligned}
& \underset{\vectory}{\text{maximize}}
& & \textbf{1}^T\vectorbeta + \sum_{h=1}^{H}\gamma_h(\bar{\vectord}^{h^T}\matrixC^h\vectory-\beta_h\|\matrixC^h\vectory\|) \\
& \text{subject to}
& & \textbf{1}^T\vectory-k=0 \\
&&& \matrixA\vectory-\vectorb \preceq \textbf{0} 
\end{aligned}
\end{equation}

If we treat $\vectorgamma$ and $\vectorbeta$ as fixed parameters, problem (4) is a convex optimization problem for decision variables $\vectory$. Indeed, the first term $\textbf{1}^T \vectorbeta$ in the objective function can be dropped since it is independent of $\vectory$,  and the second term, $\gamma_h(\bar{\vectord}^{h^T}\matrixC^h\vectory-\beta_h\|\matrixC^h\vectory\|)$, is concave in $\vectory$ for $h=1,2,\cdots, H$. 
In addition, all functions in the constraints are clearly linear. As a result, for any given $\vectorgamma$ and $\vectorbeta$, problem (4) is a convex optimization problem and can therefore be solved efficiently \citep{boyd2004convex}. However, $\vectorgamma$ and $\vectorbeta$ are not fixed parameters. How can we ensure that the solution to
problem (4) for a given pair of parameters $\vectorgamma$ and $\vectorbeta$ is indeed a solution to problem (3)? The following theorem provides clear guidance.

\textbf{Theorem 2.} Let $\vectory^*(\vectorgamma,\vectorbeta)$ be an optimal solution to problem (4) for a given pair of parameters $(\vectorgamma,\vectorbeta)$. 
It is also a stationary point solution to problem (3) if $\vectorgamma$ and $\vectorbeta$ satisfy the following two conditions:
\begin{equation}
\begin{aligned}
\beta_h\|\matrixC^h\vectory^*(\vectorgamma, \vectorbeta) \| - \bar{\vectord}^{h^T}\matrixC^h\vectory^*(\vectorgamma, \vectorbeta) =0, \ h=1, 2, \cdots, H,
\end{aligned}
\end{equation}
\vspace{-0.5cm}
\begin{equation}
\begin{aligned}
\gamma_h\|\matrixC^h\vectory^*(\vectorgamma,\vectorbeta) \|-1=0, \ h=1, 2, \cdots, H.
\end{aligned}
\end{equation}

\textbf{Proof.} See Appendix B.

Note that all optimal solutions to a continuous optimization problem are stationary points but the reverse is not always true, except for certain classes of convex optimization problems. For non-convex optimization problems such as problem (3), it is common to accept stationary points as solutions since a global optimal solution is extremely difficult to obtain, if it is possible at all \citep{rao2019engineering}.
Based on Theorem 2, we can design an algorithm to find a stationary point solution to the difficult problem (3) by iteratively solving the easy problem (4) for given parameters $(\vectorgamma,\vectorbeta)$ and updating them until conditions (5) and (6) are met with sufficient accuracy. 

We now describe the iterative algorithm. 
Consider iteration $l$ with $\vectorgamma^{l}=[\gamma_1^{l},\gamma_2^{l},\cdots,\gamma_H^{l}]^T$ and $\vectorbeta^{l}=[\beta_1^{l},\beta_2^{l},\cdots,\beta_H^{l}]^T$ being the associated parameters. Let $\vectory^l = \vectory^*(\vectorgamma^l, \vectorbeta^l)$ be an optimal solution to problem (4) for the given parameters $(\vectorgamma^l, \vectorbeta^l)$. 
To check whether conditions (5) and (6) are met with sufficient accuracy, we introduce an error vector $\boldsymbol{\delta^l} = [\delta_1^l,\cdots,\delta_H^l, \delta_{H+1}^l, \cdots, \delta_{2H}^l]^T$ and let
\begin{equation}
\begin{aligned}
\delta_h^l = \beta_h^l\|\matrixC^h\vectory^l\| - \bar{\vectord}^{h^T}\matrixC^h\vectory^l, \ h=1, 2, \cdots, H,
\end{aligned}
\end{equation}
\vspace{-0.5cm}
\begin{equation}
\begin{aligned}
\ \delta_{H+h}^l = \gamma_h^l\|\matrixC^h\vectory^l \|-1, \ h=1, 2, \cdots, H.
\end{aligned}
\end{equation}
If the error norm $ \|\boldsymbol{\delta^l}\|$ is less than a predefined threshold $\varepsilon$, indicating that conditions (5) and (6) are met with sufficient accuracy, the algorithm terminates and returns a stationary point solution $\vectory^s=\vectory^l$ to problem (3), according to Theorem 2.
Otherwise, the algorithm updates the parameters and starts a new iteration $l+1$. 
Since the parameters need to eventually meet conditions (5) and (6), we obtain the updated parameters by solving the equations defined in these conditions, giving us
\begin{equation}
\begin{aligned}
\beta_h^{l+1} = \frac{\bar{\vectord}^{h^T}\matrixC^h\vectory^{l}}{\|\matrixC^h\vectory^{l} \|}, \ h=1, 2, \cdots, H, 
\end{aligned}
\end{equation}
\vspace{-0.5cm}
\begin{equation}
\begin{aligned}
\gamma_h^{l+1} = \frac{1}{\|\matrixC^h\vectory^{l} \|}, \ h=1, 2, \cdots, H.
\end{aligned}
\end{equation}
The iterative algorithm is formally presented in Figure 1. 

{\fontfamily{lmss}
\footnotesize{
	\begin{center}
		\textbf{Figure 1 \hspace{0.1cm} An Iterative Algorithm to Solve Problem (3)}
	\end{center}	
}}
\vspace{-0.3cm}
\begin{figure}[H]
	\begin{center}
		
		\captionsetup{justification=centering,margin=2cm}
		\label{tab:table 1}
		\begin{tabular}{l} 
			\hline
			
			\textbf{Input}	\hspace{0.1cm} $\varepsilon$: convergence threshold, $\varepsilon>0$\\ 
			
			\hspace{1.2cm} $\matrixC^h$: candidate profile matrix for dimension $h$, $h=1,2,\dots,H$ \\
			
			\hspace{1.2cm} $\vectord^h$: user's diversity preference for dimension $h$, $h=1,2,\dots,H$ \\
			\textbf{Output}	$\vectory^s$: stationary point solution to problem (3)\\ 
			
			$l=1$.  \hspace{0.5cm}   //$l$: iteration count\\
		
			Randomly initialize parameters $(\vectorgamma^l, \ \vectorbeta^l)$.\\
			
			Compute $\vectory^l$ by solving problem (4) with $(\vectorgamma^l, \ \vectorbeta^l)$.\\
			
			Compute errors $\vectordelta^l$ according to Equations (7) and (8).\\
			
			\textbf{While} Not($ \|\vectordelta^l\|< \varepsilon$) \\ 
			
			$\hspace{0.6cm}$ Obtain parameters $(\vectorgamma^{l+1}, \ \vectorbeta^{l+1})$ according to Equations (9) and (10). \\
			
			$\hspace{0.75cm}$Compute $\vectory^{l+1}$ by solving problem (4) with $(\vectorgamma^{l+1}, \ \vectorbeta^{l+1})$.\\
			
			$\hspace{0.75cm}$Compute errors $\vectordelta^{l+1}$ according to Equations (7) and (8).\\
			
			$\hspace{0.75cm}$ $l=l+1$. \\
			
			\textbf{End While}\\
			
			$\vectory^s=\vectory^{l}$. \\
			Return $\vectory^s$.
			$\hspace{1cm}$\\		    
			\hline
		\end{tabular}
	\end{center}
\end{figure}

\vspace{-0.6cm}

We are now ready to propose the diversity preference-aware link recommendation (DPA-LR) method. The core component of the method is the iterative algorithm presented in Figure 1, which finds a stationary point solution $\vectory^s$ to problem (3). Recall that problem (3) is equivalent to problem (2); thus, $\vectory^s$ is also a stationary point solution to problem (2). The only difference between problem (2) and problem (1) (i.e., the DPA-LR problem) is that the decision variables in problem (2) are real numbers between 0 and 1 while those in problem (1) are either 0 or 1. To derive a solution to problem (1) from the solution for problem (2), the DPA-LR method first sorts the entries of vector $\vectory^s$ in descending order and then assigns the value 1 to the top-$k$ entries (i.e., they are recommended to the user) and 0 to the rest (i.e., they are not recommended to the user). Note that the  DPA-LR method can be run in parallel for each user, thereby speeding up the diversity preference-aware friend recommendations for users.

\section{Empirical Evaluation}
We evaluate the proposed DPA-LR method using two real-world online social network data sets. One is a public data set collected from Google+ and the other one is gathered from a major U.S. online social network. In this section, we report the evaluation results using the Google+ data set. Similar evaluation results are obtained using the other data set and are reported in Appendix D for reasons of space.

\subsection{Data and Benchmark Methods}

The Google+ data set is a public data set collected by \cite{gong2012evolution}.\footnote{The data set can be downloaded from \href{http://gonglab.pratt.duke.edu/google-dataset}{http://gonglab.pratt.duke.edu/google-dataset}.} Although Google+  was shut down in 2019, the data set features 
abundant social interactions between millions of users and is still widely used to evaluate the performance of link recommendation and social network analytics methods \citep{GongSVM, lu2016exploring, lakhotia2020gpop}. The Google+ data set used in our evaluation contains data about Google+ users' profiles as well as their linkages in July, August, and September 2011, which we denote as time periods 0, 1, and 2.\footnote{The original Google+ data set contains user profiles and linkage data for four time periods. However, not many links and users were added during the last time period, compared to the first three periods. Hence, we used data for only the first three periods in our evaluation, congruent with \citep{GongSVM}.} 
Table 2 tabulates the summary statistics of the linkage data. As shown, there are about 4.6 million users connected by around 47 million directed links in period 0, and these numbers increase to 26 million and 410 million in period 2, respectively. Links in Google+ are directed, i.e., one user follows another user. We consider that a friendship link between two users is established if there exist mutual following relationships between them, consistent with a common practice in link recommendation \citep{GongSVM, zhang2018link}. 

\begin{table}[H]
	\begin{center}
		\captionsetup{justification=centering,margin=0.5cm}
		\caption{\textbf{Summary Statistics of the Linkage Data }}
		\label{tab:table 1}
		\renewcommand{\arraystretch}{0.9}
		\begin{tabular}{c  @{\hskip 0.15in} c  @{\hskip 0.15in} c} 
			\hline
			Time Period & Number of Users & Number of Directed Links \\
			\hline
			0 & 4,693,129 & 47,130,325 \\
			1 & 17,091,929 & 271,915,755 \\
			2 & 26,244,659 & 410,445,770 \\
			\hline
		\end{tabular}
		\renewcommand{\arraystretch}{0.9}
	\end{center}
\end{table}
\vspace{-0.6cm}

In the data set, each Google+ user is described by a profile that has four dimensions: major, school, employer, and place lived. Because users self-report their profiles, the same user profile might be recorded in different ways in the data set. For example, the employer Google is recorded as \textquotedblleft google," \textquotedblleft google inc," \textquotedblleft google, inc.," \textquotedblleft google.com," etc., in the data set. To address this issue, we used a popular tool named String Grouper\footnote{A description of the tool can be found at \href{https://bergvca.github.io/2020/01/02/string-grouper.html}{\textcolor{blue}{https://bergvca.github.io/2020/01/02/string-grouper.html}}, and the latest version of the tool can be downloaded from \href{https://pypi.org/project/string-grouper/}{\textcolor{blue}{https://pypi.org/project/string-grouper/}}. This tool is commonly applied to merge similar strings, as in \cite{de2015learning}.} to group different values of a dimension for the same profile into one value for that profile. 
Table 3 reports the summary statistics of the cleaned profile data. In this table, $n_h$ denotes the number of unique values for dimension $h$. For example, there are 12,242 unique majors in the data set. The notation $max(u_h)$ and $avg(u_h)$ respectively indicate the maximum and average number of profile values a user has for dimension $h$. For example, a user has as many as 17 majors and an average of 1.27 majors in the data set.

\begin{table}[H]
	\begin{center}
		\captionsetup{justification=centering,margin=0.5cm}
		\caption{\textbf{Summary Statistics of the Cleaned Profile Data}}
		\label{tab:table 1}
		\renewcommand{\arraystretch}{0.9}
		\begin{tabular}{c @{\hskip 0.3in} c @{\hskip 0.3in} c @{\hskip 0.3in} c} 
			\hline 
			Dimension($h$) & $n_h$ &  $max(u_h)$ & $avg(u_h)$ \\ \hline
			Major & 12,242 & 17 & 1.27\\
			School & 33,930 & 24 & 1.30\\
			Employer & 24,360 & 20 & 1.24\\ 
			Place Lived & 34,827  & 86 & 2.36 \\
			\hline
		\end{tabular}
		\renewcommand{\arraystretch}{0.9}
	\end{center}
\end{table}
\vspace{-0.6cm}

Because the DPA-LR problem is a new research problem, no existing method has been developed for this problem. A closely related problem is the diversification problem in the field of recommender systems, and various diversification methods have been proposed to diversify recommendations by recommender systems \citep{castells2015novelty,wu2019recent}. Moreover, prior research has adapted diversification methods to improve the diversity of link recommendations \citep{sanz2018enhancing}. Therefore, we benchmarked our method against representative diversification methods. As reviewed in Section 2.2, one classical diversification method is the maximal marginal relevance-based method (MMR) proposed by  \cite{ziegler2005improving}, which has been widely applied to enhance the recommendation diversity of recommender systems. In addition, MMR has also been adapted for link recommendation \citep{sanz2018enhancing}. Therefore, we chose MMR as a benchmark method. We also benchmarked our method against the max-sum diversification method (MSD) \citep{zhang2008avoiding}, a prevalent diversification method, and against the determinantal point process method (DPP) \citep{chen2018fast}, a state-of-the-art diversification method. These methods diversify recommendations by balancing the trade-off between recommendation accuracy and diversity. To adapt these diversification methods for link recommendation, we measured relevance scores using linkage likelihoods and gauged recommendation diversity based on user profiles. Instead of balancing the trade-off between recommendation accuracy and diversity, \cite{boim2011diversification} propose diversifying recommendations by clustering items based on their pair-wise dissimilarities and then selecting one item with the highest relevance score from each cluster to form the final recommendation list. Therefore, we also included \cite{boim2011diversification}'s DiRec method as a benchmark. To adapt this method for link recommendation, we measured pair-wise dissimilarities between users based on their profiles and substituted linkage likelihoods for relevance scores. 
Table 4 summarizes the methods compared in our evaluation. Time complexity analysis of our method is given in Appendix C. The benchmark methods were implemented in accordance with the papers proposing them \citep{ziegler2005improving, zhang2008avoiding, chen2018fast, boim2011diversification}. For our method, we set the convergence threshold $\varepsilon$ in Figure 1 to $10^{-3}$.

\begin{table}[H]
	
	\begin{center}
		\captionsetup{justification=centering,margin=0.5cm}
		\caption{\textbf{Summary of Methods Compared in the Evaluation}}
		\label{tab:table 1}
		\renewcommand{\arraystretch}{0.9}
		\begin{tabular}{@{\hskip 0.1in}c @{\hskip 0.3in} c @{\hskip 0.1in}} 
			\hline
			Method & Note  \\
			\hline
			DPA-LR & Our proposed method\\
			MMR & Maximum Marginal Relevance, benchmark    \\ 
			MSD & Max-Sum Diversification, benchmark   \\ 
			DPP & Determinantal Point Process-based method, benchmark  \\
			DiRec & Clustering-based method, benchmark    \\ 
			\hline
		\end{tabular}
		\renewcommand{\arraystretch}{0.9}
	\end{center}
\end{table}

\vspace{-0.6cm}

\subsection{Evaluation Procedure}

We first constructed a set of candidate friends for each user, which served as the input to each compared method. Recall that a user's potential friends are those who are not friends of the user in the current time period, and a user's candidate friends are those potential friends who are most likely to befriend (link with) the user in the next time period. Thus, to construct a user's candidate friend set, we needed to predict the linkage likelihoods between the user and each of her potential friends.\footnote{Following a common practice in link recommendation \citep{backstrom2011supervised, GongSVM,li2017utility}, we focused on potential friends who are two hops away from a user. Compared to the use of all of a user's potential friends, this practice greatly improves computational efficiency without sacrificing much in terms of the accuracy of linkage likelihood prediction \citep{backstrom2011supervised,li2017utility}.} To this end, we employed a state-of-the-art link prediction method,  Graph Convolutional Network-based (GCN-based) link prediction \citep{kipf2016semi,pareja2020evolvegcn}. Each instance of the training data consisted of structural and profile information of a user and that information of one of the user's potential friends in time period 0; the instance was labeled as 1 if the user and her potential friend in the instance became friends in time period 1, or 0 otherwise. Each instance of the test data contained structural and profile information of a user and that information of one of the user's potential friends in time period 1; the test instance was labeled as 1 if the user and her potential friend in the instance became friends in time period 2, or 0 otherwise. The GCN-based link prediction method was trained using the training data to predict the linkage likelihood between a user and each of her potential friends in time period 1.\footnote{In our evaluation, GCN was implemented using Deep Graph Library (\url{https://www.dgl.ai/}), a python package for graph neural networks. The settings of GCN were congruent with those recommended in \cite{kipf2016semi} and \cite{pareja2020evolvegcn}.} Each user's top-100 potential friends with the highest linkage likelihoods were selected as her candidate friends in time period 1.

Each method (ours or a benchmark) took a user's candidate friend set in time period 1 as the input and recommended $k$ friends from that set to the user. We measure the recommendation performance as the degree to which the objective of the DPA-LR problem was satisfied, i.e., the degree to which the $k$ recommended friends met the user's diversity preference. Accordingly, we define the diversity preference matching score (DPMS) of a recommendation as the mean cosine similarity between a user's diversity preference for dimension $h$, $\vectord^h$, and the diversity of the $k$ friends recommended to the user for that dimension, $\vectorr^h$, averaged across the $H$ profile dimensions:
$$
\begin{aligned}
\text{DPMS} =\frac{1}{H}\times \sum_{h=1}^{H} \frac{\vectord^{h^T} \vectorr^h}{\|\vectord^h\| \|\vectorr^h\|}.
\end{aligned}
$$
In this equation, $\vectord^h$ was derived based on the user's existing friends before the end of time period 1 (see Example 1 for an example), and $\vectorr^h$ was constructed using the profiles of the $k$ recommended friends (see Example 3 for an example). 
The DPMS of a recommendation falls in the range $[0,1]$. The higher the value of the DPMS, the better the recommended friend meets the user's diversity preference. The DPMS of each compared method is calculated by averaging the DPMSs of its recommendations. 

In addition, we also evaluated the recommendation performance using conventional link recommendation metrics: precision, recall, and the F1 score \citep{backstrom2011supervised,li2017utility}. These metrics collectively reflect the accuracy of a recommendation. Let $TP$ denote the number of friends recommended to a user in time period 1 who actually become the user's friends in time period 2, and let $P$ be the number of the user's friends who are added in time period 2. The precision of a recommendation is the percentage of the $k$ friends recommended to a user in time period 1 who actually become the user's friends in time period 2:   
$$
\begin{aligned}
\text{precision} = \frac{TP}{k}.
\end{aligned}
$$
The recall of a recommendation is the percentage of a user's friends added in time period 2 who also appear in the set of $k$ friends recommended to the user in time period 1: 
$$
\begin{aligned}
\text{recall} = \frac{TP}{P}.
\end{aligned}
$$
The F1 score of a recommendation is the harmonic mean of precision and recall:
$$	
\begin{aligned}
\text{F1} = \frac{2 \times  \text{precision} \times  \text{recall}}{\text{precision}+\text{recall}}.
\end{aligned}
$$
The precision, recall, and F1 score for a method are computed by averaging the precisions, recalls, and F1 scores of its recommendations.

\subsection{Evaluation Results}

Following the evaluation procedure, we compared the performance of the methods in Table 4 by setting $k=10$, i.e., recommending 10 friends to a user. For the benchmark methods that balance the trade-off between recommendation accuracy and diversity, including MMR, MSD, and DPP, we set the diversity weight $\theta$ to 0.5, indicating an equal weight for recommendation accuracy and diversity.\footnote{These benchmarks aim to optimize the weighted sum of recommendation accuracy and diversity. Given the diversity weight $\theta$, the accuracy weight is $1-\theta$ \citep{ziegler2005improving, zhang2008avoiding, chen2018fast}.} Table 5 reports the performance of each method for the four evaluation metrics, with the percentage improvement of our method over a benchmark listed in parentheses under the benchmark's performance. As Table 5 shows, our method attains the best performance on all evaluation metrics. For DPMS, our method outperforms the benchmarks by a range of 117.27\% to 239.44\%, suggesting that friends recommended by DPA-LR satisfy users' diversity preferences substantially better than those recommended by any of the benchmark methods. For recommendation accuracy, our method surpasses the respective best performing benchmark by 22.17\% in precision, by 33.37\% in recall, and by 28.89\% in F1 score. Considering the sheer number of users in the data set, such improvements in recommendation accuracy mean that considerably more candidate friends recommended by our method become actual friends, compared to the benchmarks.  
We further applied the paired t-test to the performance results for the recommendations made by each method and found that our method significantly
outperformed each benchmark method across all four metrics ($p<0.001$). Overall, compared to the benchmark methods, our method not only caters better to users' diversity preferences but also provides more accurate recommendations.

\begin{table}[H]
	{\small
		\begin{center}
			\captionsetup{justification=centering,margin=0.5cm}
			\caption{\textbf{Performance Comparison between DPA-LR and Benchmarks ($k$=10)}}
			\label{tab:table 1}
			\renewcommand{\arraystretch}{0.9}
			\begin{tabular}{c@{\hskip 0.2in}c@{\hskip 0.2in}c@{\hskip 0.2in}c@{\hskip 0.2in}c} 
				\hline
				\textbf{Method}   & \textbf{DPMS} & \textbf{Precision} &  \textbf{Recall} &  \textbf{F1 Score}     \\
				\hline
				
				DPA-LR & 0.4559   & 0.1541  & 0.1559  & 0.1149  \\
				
				MMR($\theta=0.5$)   & 0.1813   & 0.1214  & 0.1089  & 0.0870  \\[-0.4em]
				& (151.43\%) & (26.95\%) & (43.11\%) & (32.07\%) \\
				MSD($\theta=0.5$)   & 0.1343   & 0.1114  & 0.1035  & 0.0810  \\[-0.4em]
				& (239.44\%) & (38.30\%) & (50.56\%) & (41.82\%) \\
				DPP($\theta=0.5$)   & 0.1968   & 0.1222  & 0.1169  & 0.0881  \\[-0.4em]
				& (131.64\%) & (26.11\%) & (33.37\%) & (30.37\%) \\
				DiRec & 0.2099   & 0.1262  & 0.1128  & 0.0892  \\[-0.4em]
				& (117.27\%) & (22.17\%) & (38.19\%) & (28.89\%) \\
				\hline
			\end{tabular}
			\renewcommand{\arraystretch}{0.9}
	
	\end{center}}
\vspace{-3mm}
\centering{\footnotesize Note: Percentage improvement of our method over a benchmark listed in parentheses.}
\end{table}

To check the robustness of our method's superiority over the benchmark methods, we conducted additional experiments by varying the number $k$ of recommended friends between 6 and 14. 
As reported in Table 6, DPA-LR substantially outperforms each benchmark on all four metrics across the investigated values of $k$. 
In particular, DPA-LR surpasses the respective best performing benchmark by a range of 96.35\% to 128.14\% in DPMS, 19.63\% to 22.75\% in precision, 27.31\% to 38.45\% in recall, and 25.87\% to 32.30\% in F1 score.  
Our method's outperformance of each benchmark is statistically significant across $k$ ($p<0.001$). 

{\fontfamily{lmss}
	\footnotesize{
		\begin{center}
			\textbf{Table 6 \hspace{0.1cm} Performance Comparison between DPA-LR and Benchmarks: $k$=6 to $k$=14}
\end{center}}}
\vspace{-4mm}

{\small
 
	\begin{longtable}{c@{\hskip 0.2in} c @{\hskip 0.2in}  c@{\hskip 0.2in} c@{\hskip 0.2in} c @{\hskip 0.2in}c@{\hskip 0.2in} c}
		
		\hline 
		\textbf{Metric} &\textbf{Method}&\textbf{$k$=6}&\textbf{$k$=8}&\textbf{$k$=10}&\textbf{$k$=12}&\textbf{$k$=14}\\
		\hline 
		\endfirsthead
		
		\multicolumn{7}{c}%
		{{\bfseries \tablename\ \thetable{} {\footnotesize{\fontfamily{lmss}\selectfont -- continued from previous page}}}} \\
		\hline 
		\textbf{Metric} &\textbf{Method}&\textbf{$k$=6}&\textbf{$k$=8}&\textbf{$k$=10}&\textbf{$k$=12}&\textbf{$k$=14}\\
		\hline 
		\endhead
		
		\hline 
		\multicolumn{7}{c}{{\footnotesize Continued on next page}} \\
		\endfoot
		\multicolumn{7}{c}{{\footnotesize Note: Percentage improvement of our method over a benchmark listed in parentheses.}} 
		\endlastfoot
		
		DPMS 		& DPA-LR & 0.3998   & 0.4100   & 0.4559   & 0.4595   & 0.4612   \\
		& MMR ($\theta=0.5$)   & 0.1596   & 0.1700   & 0.1813   & 0.1927   & 0.2038   \\[-0.4em]
		&       & (150.55\%) & (141.18\%) & (151.43\%) & (138.43\%) & (126.32\%) \\
		& MSD ($\theta=0.5$)   & 0.0885   & 0.1068   & 0.1343   & 0.1499   & 0.1645   \\[-0.4em]
		&       & (351.50\%) & (283.81\%) & (239.44\%) & (206.48\%) & (180.44\%) \\
		& DPP ($\theta=0.5$)   & 0.1455   & 0.1715   & 0.1968   & 0.2136   & 0.2303   \\[-0.4em]
		&       & (174.75\%) & (139.04\%) & (131.64\%) & (115.11\%) & (100.31\%) \\
		& DiRec & 0.1752   & 0.1943   & 0.2099   & 0.2232   & 0.2349   \\[-0.4em]
		&       & (128.14\%) & (111.06\%) & (117.27\%) & (105.88\%) & (96.35\%) \\
		\hline
		Precision   & DPA-LR & 0.1619  & 0.1577  & 0.1541  & 0.1509  & 0.1479  \\
		& MMR ($\theta=0.5$)  & 0.1274  & 0.1234  & 0.1214  & 0.1204  & 0.1193  \\[-0.4em]
		&       & (27.11\%) & (27.75\%) & (26.95\%) & (25.31\%) & (23.94\%) \\
		& MSD ($\theta=0.5$)  & 0.1099  & 0.1108  & 0.1114  & 0.1118  & 0.1117  \\[-0.4em]
		&       & (47.27\%) & (42.31\%) & (38.30\%) & (35.00\%) & (32.40\%) \\
		& DPP ($\theta=0.5$)  & 0.1277  & 0.1252  & 0.1222  & 0.1203  & 0.1193  \\[-0.4em]
		&       & (26.82\%) & (25.96\%) & (26.11\%) & (25.38\%) & (23.99\%) \\
		& DiRec & 0.1353  & 0.1299  & 0.1262  & 0.1231  & 0.1205  \\[-0.4em]
		&       & (19.63\%) & (21.44\%) & (22.17\%) & (22.52\%) & (22.75\%) \\
		\hline
		Recall 		& DPA-LR & 0.1024  & 0.1302  & 0.1559  & 0.1792  & 0.2015  \\
		& MMR ($\theta=0.5$)   & 0.0692  & 0.0889  & 0.1089  & 0.1297  & 0.1502  \\[-0.4em]
		&       & (47.97\%) & (46.47\%) & (43.11\%) & (38.23\%) & (34.18\%) \\
		& MSD ($\theta=0.5$)  & 0.0611  & 0.0821  & 0.1035  & 0.1246  & 0.1457  \\[-0.4em]
		&       & (67.45\%) & (58.57\%) & (50.56\%) & (43.80\%) & (38.35\%) \\
		& DPP ($\theta=0.5$)  & 0.0739  & 0.0961  & 0.1169  & 0.1376  & 0.1583  \\[-0.4em]
		&       & (38.45\%) & (35.48\%) & (33.37\%) & (30.22\%) & (27.31\%) \\
		& DiRec & 0.0707  & 0.0914  & 0.1128  & 0.1334  & 0.1535  \\[-0.4em]
		&       & (44.71\%) & (42.45\%) & (38.19\%) & (34.32\%) & (31.31\%) \\
		\hline
		F1 Score 			& DPA-LR & 0.0945  & 0.1061  & 0.1149  & 0.1215  & 0.1268  \\
		& MMR ($\theta=0.5$)  & 0.0693  & 0.0787  & 0.0870  & 0.0945  & 0.1007  \\[-0.4em]
		&       & (36.37\%) & (34.79\%) & (32.07\%) & (28.51\%) & (25.87\%) \\
		& MSD ($\theta=0.5$)  & 0.0603  & 0.0715  & 0.0810  & 0.0889  & 0.0954  \\[-0.4em]
		&       & (56.69\%) & (48.39\%) & (41.82\%) & (36.66\%) & (32.84\%) \\
		& DPP ($\theta=0.5$)  & 0.0705  & 0.0807  & 0.0881  & 0.0946  & 0.1004  \\[-0.4em]
		&       & (34.05\%) & (31.55\%) & (30.37\%) & (28.45\%) & (26.28\%) \\
		& DiRec & 0.0714  & 0.0810  & 0.0892  & 0.0955  & 0.1005  \\[-0.4em]
		&       & (32.30\%) & (31.00\%) & (28.89\%) & (27.22\%) & (26.07\%) \\
		\hline

\end{longtable}}

\vspace{-0.2cm}

We performed additional robustness checks by changing the diversity weight $\theta$ for the benchmarks MMR, MSD, and DPP from 0.1 to 0.9 with an increment of 0.1. As $\theta$ increases from 0.1 to 0.9, these benchmarks weight diversity more and recommendation accuracy less, and they change from highly skewed toward accuracy ($\theta=0.1$) to balanced accuracy and diversity ($\theta=0.5$) and then to highly skewed toward diversity ($\theta=0.9$). As reported in Table 7, our method substantially and significantly outperforms each benchmark across the values of $\theta$ ($p<0.001$). 
We also compared the performance of our method and the benchmarks using another large-scale online social network data set and report the superior performance of our method over the benchmarks in Appendix D. Taken together, our evaluations demonstrate the superiority of our method over the benchmarks across different numbers $k$ of recommended friends, with various diversity weights $\theta$, and using different data sets.
\vspace{0.56cm}
{\fontfamily{lmss}
	\footnotesize{
		\begin{center}
			\textbf{Table 7 \hspace{0.1cm} Performance Comparison between DPA-LR and Benchmarks ($k$=10): $\theta=0.1$ to $\theta=0.9$}
\end{center}}}
\vspace{-4mm}

{\small
	
	\begin{longtable}{c@{\hskip 0.18in} c@{\hskip 0.2in} c @{\hskip 0.2in}c @{\hskip 0.2in}c @{\hskip 0.2in}c }
		
		\hline 
		
		\textbf{Method} &  \textbf{Diversity Weight}  & \textbf{DPMS} & \textbf{Precision} & \textbf{Recall} & \textbf{F1 Score}  \\
		\hline 
		\endfirsthead
		
		\multicolumn{6}{c}%
		{{\bfseries \tablename\ \thetable{} {\footnotesize\fontfamily{lmss}\selectfont -- continued from previous page}}} \\
		\hline 
		\textbf{Method} &  \textbf{Diversity Weight}  & \textbf{DPMS} & \textbf{Precision} & \textbf{Recall} & \textbf{F1 Score} \\
		\hline 
		\endhead
		
		\hline 
		\multicolumn{6}{c}{{\footnotesize Continued on next page}} \\
		\endfoot
		\multicolumn{6}{c}{{{\footnotesize Note: Percentage improvement of our method over a benchmark listed in parentheses.}}} 
		
		\endlastfoot

		DPA-LR   & -       & 0.4559   & 0.1541  & 0.1559  & 0.1149  \\
		\hline
		MMR     & $\theta=0.1$ & 0.2265   & 0.1366  & 0.1152  & 0.0953  \\[-0.4em]
		&              & (101.27\%) & (12.85\%) & (35.34\%) & (20.54\%) \\
		& $\theta=0.2$ & 0.2086   & 0.1327  & 0.1142  & 0.0935  \\[-0.4em]
		&              & (118.56\%) & (16.16\%) & (36.54\%) & (22.90\%) \\
		& $\theta=0.3$ & 0.1948   & 0.1284  & 0.1121  & 0.0910  \\[-0.4em]
		&              & (134.03\%) & (20.05\%) & (39.06\%) & (26.23\%) \\
		& $\theta=0.4$ & 0.1866   & 0.1246  & 0.1103  & 0.0889  \\[-0.4em]
		&              & (144.34\%) & (23.73\%) & (41.39\%) & (29.30\%) \\
		& $\theta=0.5$ & 0.1813   & 0.1214  & 0.1089  & 0.0870  \\[-0.4em]
		&              & (151.43\%) & (26.95\%) & (43.11\%) & (32.07\%) \\
		& $\theta=0.6$ & 0.1557   & 0.1022  & 0.0936  & 0.0740  \\[-0.4em]
		&              & (192.78\%) & (50.78\%) & (66.52\%) & (55.25\%) \\
		& $\theta=0.7$ & 0.1745   & 0.1187  & 0.1101  & 0.0860  \\[-0.4em]
		&              & (161.28\%) & (29.88\%) & (41.62\%) & (33.63\%) \\
		& $\theta=0.8$ & 0.1724   & 0.1186  & 0.1131  & 0.0866  \\[-0.4em]
		&              & (164.44\%) & (30.01\%) & (37.83\%) & (32.68\%) \\
		& $\theta=0.9$ & 0.1719   & 0.1191  & 0.1189  & 0.0883  \\[-0.4em]
		&              & (165.17\%) & (29.38\%) & (31.09\%) & (30.08\%) \\
		
		\hline
		MSD 	& $\theta=0.1$ & 0.1960   & 0.1282  & 0.1109  & 0.0906  \\[-0.4em]
		&              & (132.66\%) & (20.23\%) & (40.53\%) & (26.80\%) \\
		& $\theta=0.2$ & 0.1756   & 0.1229  & 0.1083  & 0.0875  \\[-0.4em]
		&              & (159.60\%) & (25.44\%) & (43.97\%) & (31.25\%) \\
		& $\theta=0.3$ & 0.1601   & 0.1188  & 0.1064  & 0.0852  \\[-0.4em]
		&              & (184.73\%) & (29.71\%) & (46.51\%) & (34.91\%) \\
		& $\theta=0.4$ & 0.1467   & 0.1149  & 0.1044  & 0.0828  \\[-0.4em]
		&              & (210.80\%) & (34.18\%) & (49.33\%) & (38.74\%) \\
		& $\theta=0.5$ & 0.1343   & 0.1114  & 0.1035  & 0.0810  \\[-0.4em]
		&              & (239.44\%) & (38.30\%) & (50.56\%) & (41.82\%) \\
		& $\theta=0.6$ & 0.1230   & 0.1079  & 0.1024  & 0.0790  \\[-0.4em]
		&              & (270.77\%) & (42.82\%) & (52.31\%) & (45.41\%) \\
		& $\theta=0.7$ & 0.1129   & 0.1047  & 0.1018  & 0.0773  \\[-0.4em]
		&              & (303.91\%) & (47.19\%) & (53.15\%) & (48.61\%) \\
		& $\theta=0.8$ & 0.1052   & 0.1017  & 0.1014  & 0.0757  \\[-0.4em]
		&              & (333.57\%) & (51.58\%) & (53.72\%) & (51.88\%) \\
		& $\theta=0.9$ & 0.1008   & 0.1000  & 0.1010  & 0.0749  \\[-0.4em]
		&              & (352.15\%) & (54.21\%) & (54.33\%) & (53.36\%) \\
		
		\hline
		DPP 	& $\theta=0.1$ & 0.1969   & 0.1228  & 0.1177  & 0.0886  \\[-0.4em]
		&              & (131.56\%) & (25.50\%) & (32.42\%) & (29.73\%) \\
		& $\theta=0.2$ & 0.1970   & 0.1227  & 0.1176  & 0.0885  \\[-0.4em]
		&              & (131.43\%) & (25.60\%) & (32.55\%) & (29.83\%) \\
		& $\theta=0.3$ & 0.1969   & 0.1226  & 0.1174  & 0.0884  \\[-0.4em]
		&              & (131.56\%) & (25.72\%) & (32.75\%) & (29.96\%) \\
		& $\theta=0.4$ & 0.1968   & 0.1224  & 0.1172  & 0.0883  \\[-0.4em]
		&              & (131.64\%) & (25.89\%) & (33.01\%) & (30.14\%) \\
		& $\theta=0.5$ & 0.1968   & 0.1222  & 0.1169  & 0.0881  \\[-0.4em]
		&              & (131.64\%) & (26.11\%) & (33.37\%) & (30.37\%) \\
		& $\theta=0.6$ & 0.1968   & 0.1219  & 0.1165  & 0.0879  \\[-0.4em]
		&              & (131.66\%) & (26.42\%) & (33.87\%) & (30.72\%) \\
		& $\theta=0.7$ & 0.1968   & 0.1214  & 0.1157  & 0.0875  \\[-0.4em]
		&              & (131.69\%) & (26.95\%) & (34.79\%) & (31.35\%) \\
		& $\theta=0.8$ & 0.1968   & 0.1205  & 0.1141  & 0.0867  \\[-0.4em]
		&              & (131.70\%) & (27.94\%) & (36.64\%) & (32.50\%) \\
		& $\theta=0.9$ & 0.1968   & 0.1182  & 0.1106  & 0.0848  \\[-0.4em]
		&              & (131.71\%) & (30.36\%) & (40.95\%) & (35.42\%) \\
		\hline

\end{longtable}}
\vspace{-0.2cm}

\subsection{Analysis of Performance}

Having demonstrated the superiority of our DPA-LR method over the benchmarks, it is intriguing to connect its superior performance to its methodological novelty. Recall that our method takes candidate friends produced by GCN-based link prediction as inputs and chooses which of these candidate friends to recommend by optimizing each user's diversity preference at the dimension level (this is its methodological novelty). If we drop this methodological novelty, our method reduces to the method that recommends friends with the highest linkage likelihoods predicted by GCN-based link prediction, namely GCN-based link recommendation (GCN-LR). We implemented GCN-LR and compared its performance with that of our method. As reported in Table 8, compared to GCN-LR, our method improves DPMS by 90.89\% ($p<0.001$), which can be attributed to its optimization of each user's diversity preference. Moreover, because friends recommended by our method better satisfy a user's diversity preference (i.e., an improved DPMS), they are more likely to be accepted by the user. Consequently, our method substantially and significantly outperforms GCN-LR in precision, recall, and F1 score ($p<0.001$). These performance improvements are directly contributed by the methodological novelty of our method, as our method reduces to GCN-LR by dropping the methodological novelty. We also report the performance of a representative benchmark method (MMR) in Table 8. Like our method, MMR also takes candidate friends produced by GCN-based link prediction as inputs. Different from our method, however, MMR increases the diversity of friends recommended to a user at the cost of recommendation accuracy \citep{kaminskas2016diversity}. As a result, MMR performs worse than GCN-LR in precision, recall, and the F1 score (i.e., recommendation accuracy) as well as DPMS (as diversity is different from diversity preference). Taken together, the methodological novelty of our method contributes to its outperformance of GCN-LR, while the design of the benchmark methods leads to their underperformance with respect to GCN-LR, which explains the superiority of our method over the benchmarks.

\setcounter{table}{7} \renewcommand{\thetable}{\arabic{table}}

\begin{table}[H]
	{\small
		\begin{center}
			\captionsetup{justification=centering,margin=0.5cm}
			\caption{\textbf{Performance Comparison between DPA-LR, GCN-LR, and MMR ($k$=10)}}
			\label{tab:table 1}
			\renewcommand{\arraystretch}{0.9}
			\begin{tabular}{c@{\hskip 0.2in}c@{\hskip 0.2in}c@{\hskip 0.2in}c@{\hskip 0.2in}c} 
				\hline
				\textbf{Method}   & \textbf{DPMS} & \textbf{Precision} &  \textbf{Recall} &  \textbf{F1 Score}     \\
				\hline
				
				DPA-LR & 0.4559   & 0.1541  & 0.1559  & 0.1149  \\
				
				GCN-LR 	& 0.2389  & 0.1380  & 0.1147  & 0.0956  \\[-0.4em]
				& (90.89\%) & (11.72\%) & (35.98\%) & (20.24\%) \\
				
				MMR ($\theta=0.5$) & 0.1813   & 0.1214  & 0.1089  & 0.0870  \\[-0.4em]
				& (151.43\%) & (26.95\%) & (43.11\%) & (32.07\%) \\				
				\hline
			\end{tabular}
			\renewcommand{\arraystretch}{0.9}
	\end{center}}
\vspace{-3mm}
\centering{\footnotesize Note: Percentage improvement of our method over a benchmark listed in parentheses.}
\end{table}
\vspace{-2mm}

To further scrutinize the superior performance of our method, we modified benchmark method MMR so that it aims to meet each user's diversity preference, where diversity preference matching is formulated the same as in the objective function of the DPA-LR problem (i.e., problem (1)). The modified method, namely, diversity preference-aware MMR (DPA-MMR), maximizes the weighted sum of diversity preference matching and recommendation accuracy, whereas MMR optimizes the weighted sum of diversity and recommendation accuracy. We implemented DPA-MMR with a range of weights $\sigma$ for diversity preference matching, i.e., $\sigma=0.1, 0.5, 0.9$. Table 9 compares the performance of our method and DPA-MMR. As reported in Table 9, our method substantially and significantly outperforms DPA-MMR across $\sigma$ on all performance measures ($p<0.001$). Since both our method and DPA-MMR consider diversity preference, the superiority of our method over DPA-MMR can be solely attributed to our method's methodological advantage.

\begin{table}[H]
	{\small
		\begin{center}
			\captionsetup{justification=centering,margin=0.5cm}
			\caption{\textbf{Performance Comparison between DPA-LR and DPA-MMR ($k$=10)}}
			\label{tab:table 1}
			\renewcommand{\arraystretch}{0.9}
			\begin{tabular}{c@{\hskip 0.2in}c@{\hskip 0.2in}c@{\hskip 0.2in}c@{\hskip 0.2in}c} 
				\hline
				\textbf{Method}   & \textbf{DPMS} & \textbf{Precision} &  \textbf{Recall} &  \textbf{F1 Score}     \\
				\hline
				
				DPA-LR                & 0.4559  & 0.1541 & 0.1559  & 0.1149  \\
				DPA-MMR ($\sigma=0.1$) & 0.2332  & 0.1415 & 0.1179  & 0.0979  \\[-0.4em]
				& (95.56\%) & (8.96\%) & (32.29\%) & (17.32\%) \\
				DPA-MMR ($\sigma=0.5$) & 0.3096  & 0.1420 & 0.1255  & 0.1007  \\[-0.4em]
				& (47.26\%) & (8.56\%) & (24.28\%) & (14.13\%) \\
				DPA-MMR ($\sigma=0.9$) & 0.3696  & 0.1415 & 0.1395  & 0.1044  \\[-0.4em]
				& (23.36\%) & (8.91\%) & (11.79\%) & (10.06\%) \\
				\hline

			\end{tabular}
			\renewcommand{\arraystretch}{0.9}
	\end{center}}
\vspace{-3mm}
\centering{\footnotesize Note: Percentage improvement of our method over a benchmark listed in parentheses.}
\end{table}
\vspace{-2mm}

It is also interesting to compare the performance of our method with that of state-of-the-art link recommendation methods, which apply graph neural networks to represent users' structural and profile information, feed these representations into a classifier (e.g., MLP) to predict linkage likelihoods, and recommend friends with the highest linkage likelihoods \citep{wu2020comprehensive}. In particular, we implemented three state-of-the-art link recommendation methods, each of which was developed based on a representative graph neural network: graph convolutional network-based link recommendation \cite[GCN-LR]{kipf2016semi}, GraphSage-based link recommendation \citep[GraphSage-LR]{hamilton2017inductive}, and graph attention network-based link recommendation \citep[GAT-LR]{velivckovic2017graph}.\footnote{GCN, GraphSage, and GAT were implemented using Deep Graph Library (\url{https://www.dgl.ai/}) with settings recommended in their respective papers.} Table 10 compares the performance between these methods and our DPA-LR method for $k=10$. As reported, DPA-LR substantially and significantly outperforms each of the compared methods on all performance metrics ($p<0.001$). The superiority of our method is attributed to its consideration of diversity preference, which is neglected by existing link recommendation methods. Consequently, friends recommended by our method meet a user's diversity preference better (better DPMS) and thus are more likely to be accepted by the user (higher precision, recall and F1 score). We further varied $k$ between 6 and 14 and obtained similar results, as reported in Appendix E.

\begin{table}[H]
	{\small
		\begin{center}
			\captionsetup{justification=centering,margin=0.5cm}
			\caption{\textbf{Comparison between DPA-LR and State-of-the-art Link Recommendation Methods ($k$=10)}}
			\label{tab:table 1}
			\renewcommand{\arraystretch}{0.9}
			\begin{tabular}{c@{\hskip 0.2in}c@{\hskip 0.2in}c@{\hskip 0.2in}c@{\hskip 0.2in}c} 
				\hline
				\textbf{Method}   & \textbf{DPMS} & \textbf{Precision} &  \textbf{Recall} &  \textbf{F1 Score}     \\
				\hline
				
				DPA-LR & 0.4559   & 0.1541  & 0.1559  & 0.1149  \\
				
				GCN-LR 	& 0.2389  & 0.1380  & 0.1147  & 0.0956  \\[-0.4em]
				& (90.89\%) & (11.72\%) & (35.98\%) & (20.24\%) \\
				
				GraphSage-LR & 0.1895   & 0.1234  & 0.1341  & 0.0950  \\[-0.4em]
				& (140.59\%) & (24.94\%) & (16.26\%) & (20.95\%) \\
				
				GAT-LR       & 0.1992   & 0.1206  & 0.1194  & 0.0890  \\[-0.4em]
				& (128.89\%) & (27.79\%) & (30.61\%) & (29.16\%) \\
				
				\hline
			\end{tabular}
			\renewcommand{\arraystretch}{0.9}
	\end{center}}
\vspace{-3mm}
\centering{\footnotesize Note: Percentage improvement of our method over a benchmark listed in parentheses.}
\end{table}

\vspace{-2mm}

We also analyzed the convergence of the iterative algorithm presented in Figure 1, the core component of our method. Figure 2 plots the error norm $\|\vectordelta^l\|$ after each iteration for a user. As shown, the algorithm converges after 8 iterations, i.e., $ \|\vectordelta^l\|< \varepsilon$, where the convergence threshold $\varepsilon$ is set to $10^{-3}$. Averaged across the users, the algorithm converges after 7 iterations. Additionally, we observed that the parameters $(\vectorgamma, \vectorbeta)$ converged to the same solution irrespective of their initializations. Figure 3 plots the convergence of $(\vectorgamma, \vectorbeta)$ for a user, with 1,000 different random initializations.\footnote{For the Google+ data set, both $\vectorgamma$ and $\vectorbeta$ have four dimensions: major, school, employee, and place lived.} As shown, these parameters converge to the same solution, regardless of their initializations. Recall that the output of the iterative algorithm, i.e., the stationary point solution for problem (3), is derived by solving problem (4) with the parameters $(\vectorgamma, \vectorbeta)$. Therefore, the algorithm produces the same stationary point solution, independent of the initializations of $(\vectorgamma, \vectorbeta)$.     

{\fontfamily{lmss}
\footnotesize{
\begin{center}
	\textbf{Figure 2 \hspace{0.1cm} Convergence of the Iterative Algorithm}
\end{center}}}
\vspace{-2mm}
\begin{figure}[H]
	\centering
	\includegraphics[width=10cm]{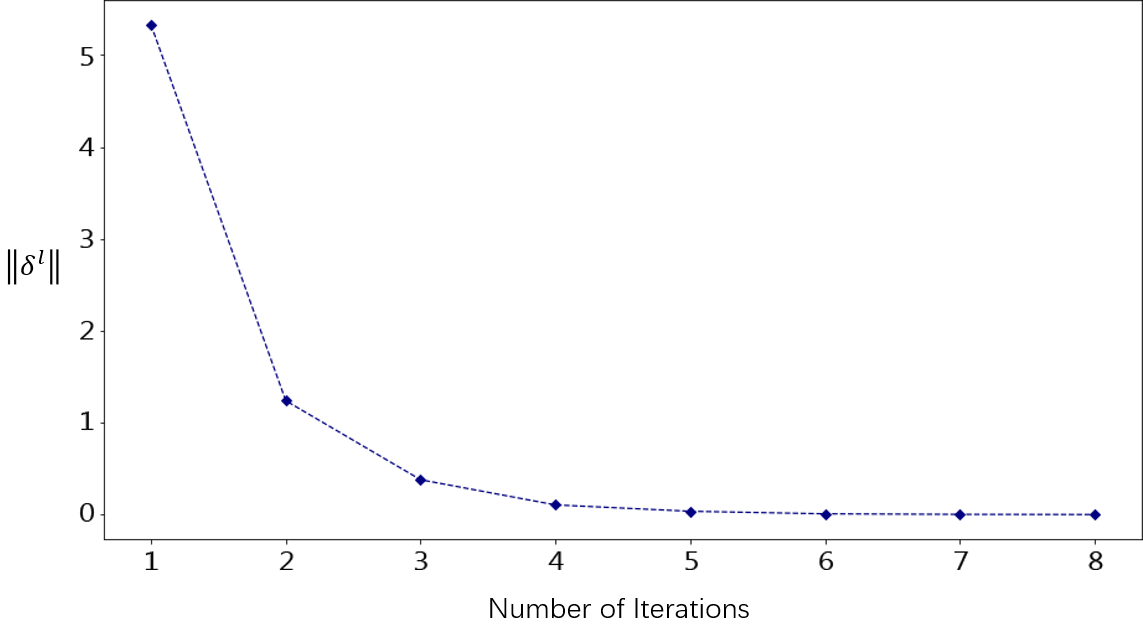}
\end{figure}

\newpage

{\fontfamily{lmss}
	\footnotesize{
		\begin{center}
			\textbf{Figure 3 \hspace{0.1cm} Parameter Convergence with 1,000 Different Random Initializations}
\end{center}}}
\vspace{-2mm}
\begin{figure}[H]
	\centering
	\includegraphics[width=16.5cm]{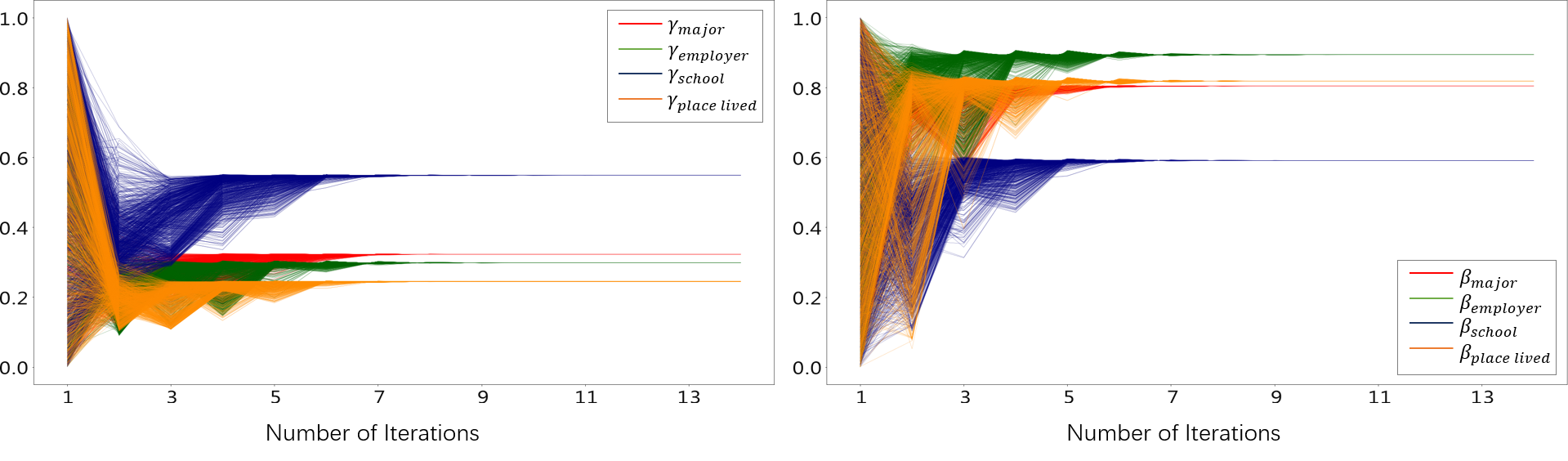}
\end{figure}
	
Finally, it is interesting to analyze the gap between the approximate solution derived by our method and the optimal solution to the DPA-LR problem (i.e., problem (1)). Recall that problem (1) is a combinatorial optimization problem of selecting $k$ friends out of $|\mathbb{C}^i|$ candidate friends for a user. We derive the optimal solution using exhaustive search, which goes through all possible combinations of $k$ candidate friends and finds the one that optimizes the objective function of problem (1). Conducting exhaustive search for large $|\mathbb{C}^i|$ is computational prohibitive. Therefore, we set $|\mathbb{C}^i|$ to 
30, 40, and 50, respectively, and $k$ to 5. The number of possible combinations of selecting $k=5$ friends out of 30, 40, and 50 candidates are 142506, 658008, and 2118760, respectively. We evaluate the gap between the approximate solution and the optimal solution using the following two metrics. First, we measure the difference between the objective value attained by the approximate solution and that attained by the optimal solution, where an objective value is obtained by evaluating the objective function of problem (1) with a solution. We call this metric ``objective difference''. Second, we measure the overlap between the friends recommended by the approximate solution and those recommended by the optimal solution. We name this metric ``recommendation overlap''. 

We ran experiments for 100 randomly sampled users by recommending $k=5$ friends out of $|\mathbb{C}^i|$ candidates for each user. As reported in Table 11, when $|\mathbb{C}^i|=30$, the average objective value attained by the optimal solution across the 100 users is 1.9382 whereas the average objective value attained by the approximate solution is 1.9012; and the objective difference between the two solutions is 1.91\%, which means that the average objective value attained by the approximate solution is 1.91\% lower than that attained by the optimal solution. We also observe small objective differences for the cases of $|\mathbb{C}^i|=40$ and $|\mathbb{C}^i|=50$. For the other metric, the average recommendation overlap is 4.17 when $|\mathbb{C}^i|=30$. On average, out of 5 recommended friends, 4.17 friends recommended by the approximate solution are the same as those recommended by the optimal solution. We also observe large recommendation overlaps for the cases of $|\mathbb{C}^i|=40$ and $|\mathbb{C}^i|=50$. Overall, as evidenced by small objective differences and large recommendation overlaps in Table 11, the approximate solution by our method is a close approximation of the optimal solution. We conducted additional analyses regarding the ranking quality of our method as well as its sensitivity to the number of candidate friends in Appendix F. 

\begin{table}[H]
	{\small
	\begin{center}
		\captionsetup{justification=centering,margin=0.5cm}
		\caption{\textbf{Empirical Analysis of Gap Between Approximate Solution and Optimal Solution ($k=5$)}}
		\label{tab:table 1}
		\renewcommand{\arraystretch}{0.9}
		\begin{tabular}{@{\hskip 0.1in}c @{\hskip 0.1in}| c @{\hskip 0.1in}| c  @{\hskip 0.1in} | c @{\hskip 0.1in}  } 
			\hline 
			\textbf{Number of } &\textbf{Solution} & \textbf{Objective}  &  \textbf{Recommendation} \\
			\textbf{Candidate Friends $|\mathbb{C}^i|$ } &  & \textbf{Difference}  &  \textbf{Overlap} \\
			\hline
			                       & Optimal & 1.9382 &  \\
			  $|\mathbb{C}^i|=30$  & Approximate & 1.9012 &  4.17  \\[-0.4em]
			                       &       & (1.91\%) &   \\
			\hline
			                       & Optimal &  2.0109 &  \\
			 $|\mathbb{C}^i|=40$   & Approximate & 1.9816  & 4.27   \\[-0.4em]
        			 			   &    & (1.46\%)  &  \\
			\hline
			                       & Optimal & 2.1911  &  \\
			 $|\mathbb{C}^i|=50$   & Approximate & 2.1395  &  4.12 \\[-0.4em]
								   &    & (2.35\%)  &  \\
			\hline
		\end{tabular}
		\renewcommand{\arraystretch}{0.9}
	\end{center}}
	\vspace{-0.2cm}
\end{table}

\section{Discussion and Conclusion}

Link recommendation is a fundamental social network analytics problem with ample business implications. Existing link recommendation methods overlook diversity preference, although social psychology theories suggest its critical role for link recommendation performance \citep{burt1998personality, rivera2010dynamics, laakasuo2017company}. To address this limitation, we define and operationalize the concept of diversity preference for link recommendation and propose a new link recommendation problem, the diversity preference-aware link recommendation problem. We then develop a novel link recommendation method to solve the problem. Using two large-scale online social network data sets, we demonstrate the superior performance of our method over prevalent existing methods through extensive empirical evaluations. 

Our study belongs to the computational genre of design science research that develops computational algorithms and methods to solve business and societal problems \citep{gupta_traits_2018,padmanabhan2022machine}. According to the knowledge contribution framework of design science research \citep{gregor2013positioning}, 
our study contributes to extant literature in both problem formulation and method development. First, the proposed diversity preference-aware link recommendation problem is new. It is distinct from existing link recommendation problems in its consideration and operationalization of diversity preference. It also differs from diversification problems for recommender systems because of the difference between diversity preference researched in our problem and diversity studied in diversification problems. Second, the new link recommendation problem demands an innovative solution method. Thus, the second contribution of our study is the developed diversity preference-aware link recommendation method. Specifically, we analyze key properties of the new link recommendation problem in Theorems 1 and 2 and then develop a novel link recommendation method that optimizes each individual
user’s diversity preference at the dimension level. These theorems and the method together constitute the methodological contribution of our study.

Our study offers several implications for research. First, it informs the research community about a new perspective on link recommendation---diversity preference-aware link recommendation. Social psychology theories stipulate that friendship formations on social networks are driven by the combined effects of homophily (i.e., a user's tendency to connect with those similar to her) and heterophily (i.e., a user's tendency to connect with those complementary to her), and different users weight homophily and heterophily differently, thereby exhibiting different diversity preferences \citep{burt1998personality, rivera2010dynamics, laakasuo2017company}. Existing link recommendation methods rely solely on homophily, whereas diversification methods adapted from recommender systems focus on heterophily. Consequently, neither of these methods can recommend friends that cater well to a user's diversity preference. Unlike these methods, our method recommends friends to satisfy a user's diversity preference. Therefore, friends recommended by our method not only better meet a user's diversity preference, but also are more likely to be accepted by the user. Moreover, a user's diversity preference varies across different profile dimensions \citep{rivera2010dynamics}. Accordingly, our method is designed to optimize each user's diversity preference at the dimension level. Future research could adopt this prescription and consider individual user's diversity preference at the dimension level for more effective link recommendations. Second, our study has implications for the design of recommender systems \citep{adomavicius2005toward,he2019mobile}. Classical accuracy-maximizing recommender systems tend to recommend items that are similar to each other. Diversification methods for recommender systems improve the diversity of the recommended items but suffer from two limitations \citep{ziegler2005improving,chen2018fast}. These methods indiscriminately increase the diversity of the recommended items for every user, although not every user prefers more diversity. In addition, these methods sacrifice recommendation accuracy for diversity \citep{kaminskas2016diversity}. Informed by our study, future research could operationalize diversity preference for recommender systems and optimize this for each user at the dimension level. In doing so, items that are recommended will meet each individual user's diversity preference and will thus be more likely to be adopted by the user, thereby addressing the two limitations associated with diversification methods. Third, we model the diversity preference-aware link recommendation problem as a nonlinear sum-of-ratios binary integer programming problem (i.e., problem (1)), show the NP-hardness of the problem, and solve it with a novel iterative method. Our model is general, and there are other important problems that can be modeled as problem (1), e.g., the energy resource allocation problem \citep{boshkovska2015practical}. As a result, our proposed method can be extended to solve these important problems as well.  

This study also provides implications for business. We show that our method outperforms state-of-the-art link recommendation practices in both satisfying users' diversity preferences and recommendation accuracy. Improvements in recommendation accuracy lead to a more densely connected online social network, which in turn better facilitates information diffusion in the network and generates more revenue (e.g., advertisement revenue) for the network operator \citep{shriver2013social}. Moreover, meeting a user's diversity preference enhances the user's experience, satisfaction, and engagement with the online social network, motivates the user to spend more time using the network, and thus brings more revenue to the network operator \citep{chen2009make, qiu2017understanding}. Therefore, online social network operators should consider incorporating diversity preference into the link recommendation functionality for improved user experiences and revenue. In addition, our study informs online social network operators that diversity preference should be measured and optimized at the dimension level, as the same user could have different degrees of diversity preference for different profile dimensions. Finally, our proposed diversity preference measure is generic and can be adapted to measure users' diversity preferences in different real-world online social networks. For example, it can be applied to measure users' diversity preferences in Instagram based on their demographical and behavioral profiles (e.g., tags). Likewise, our proposed diversity preference-aware link recommendation method is also general. It can easily be integrated with a link recommendation component of an online social network, by taking linkage likelihoods predicted by the component as inputs and producing diversity preference-aware friend recommendations as outputs. 

Our study can be extended in several ways. First, our study treats diversity preference for each profile dimension as equally important. Future work should account for situations where diversity preferences for different dimensions are weighted differently. To this end, the objective function of problem (1) needs to be changed to a weighted sum of cosine similarities between a user's diversity preference and the diversity distribution of recommended friends across all dimensions. Note that our diversity preference-aware link recommendation method can still solve the updated problem (1). Second, the diversity preference measure can be extended to capture diversity preference across dimensions. For example, a user may prefer to befriend those with IS major and from New York. Toward that end, we could create a super-dimension as a combination of interested profile dimensions, e.g., major and city, and measure a user's diversity preference over the super-dimension. Our proposed method is still applicable to this extension. Third, a user's diversity preference might change over time. Therefore, another area that merits future exploration is to design a link recommendation method that considers the dynamics of diversity preference. Prior literature on data stream mining and knowledge refreshing would inform the design of this method \citep{fang2013right}. Fourth, it would be interesting to extend our study to recommender systems. To this end, future work needs to define and measure diversity preference in the context of recommender systems and design a diversity preference-aware method for recommender systems. An attempt in this direction is given in Appendix G. Fifth, we evaluate diversity preference and link recommendation methods using archival data. Future work should consider employing experiments and surveys in the evaluation. For example, we could use surveys to assess to what extent archival data can capture diversity preference and combine surveys and archival data to better model diversity preference. It is also worthwhile to conduct field or lab experiments to compare the performance between our method and benchmarks. A well-designed experiment could control factors that would bias the comparison results. In addition, in an experiment, we can observe how users react to recommended friends and which recommended friends they actually connect with in real time. Finally, conventional accuracy-oriented measures for link recommendation, such as precision and recall, are defined based on whether a recommended friend is accepted by a user. New measures gauged with the intensity of interactions (e.g., comments and likes) between a user and a friend recommended to the user could be considered. Understandably, an effective friend recommendation that caters a user's diversity preference is more likely to be accepted by the user and results in more interactions between the recommended friend and the user if they actually become friends. By evaluating a link recommendation method with both accuracy-oriented measures and these new measures, we can have a fuller understanding of the effectiveness of the method. 

\bibliographystyle{informs2014} 
\bibliography{DPA-LP} 

\begin{thebibliography}{8}
\providecommand{\natexlab}[1]{#1}
\providecommand{\url}[1]{\texttt{#1}}
\expandafter\ifx\csname urlstyle\endcsname\relax
  \providecommand{\doi}[1]{doi: #1}\else
  \providecommand{\doi}{doi: \begingroup \urlstyle{rm}\Url}\fi

\bibitem[Boyd et~al.(2004)Boyd, Boyd, and Vandenberghe]{boyd2004convex}
S.~Boyd, S.~P. Boyd, and L.~Vandenberghe.
\newblock \emph{Convex Optimization}.
\newblock Cambridge university press, 2004.

\bibitem[Bubeck(2014)]{bubeck2014convex}
S.~Bubeck.
\newblock Convex optimization: Algorithms and complexity.
\newblock \emph{arXiv preprint arXiv:1405.4980}, 2014.

\bibitem[Felfernig et~al.(2018)Felfernig, Boratto, Stettinger, and
  Tkal{\v{c}}i{\v{c}}]{felfernig2018evaluating}
A.~Felfernig, L.~Boratto, M.~Stettinger, and M.~Tkal{\v{c}}i{\v{c}}.
\newblock Evaluating group recommender systems.
\newblock In \emph{Group recommender systems}, pages 59--71. Springer, 2018.

\bibitem[Griva et~al.(2009)Griva, Nash, and Sofer]{griva2009linear}
I.~Griva, S.~G. Nash, and A.~Sofer.
\newblock \emph{Linear and nonlinear optimization}, volume 108.
\newblock Siam, 2009.

\bibitem[He et~al.(2019)He, Fang, Liu, and Li]{he2019mobile}
J.~He, X.~Fang, H.~Liu, and X.~Li.
\newblock Mobile app recommendation: an involvement-enhanced approach.
\newblock \emph{MIS Quarterly}, 43\penalty0 (3):\penalty0 827--849, 2019.

\bibitem[He et~al.(2017)He, Liao, Zhang, Nie, Hu, and Chua]{he2017neural}
X.~He, L.~Liao, H.~Zhang, L.~Nie, X.~Hu, and T.-S. Chua.
\newblock Neural collaborative filtering.
\newblock In \emph{Proceedings of the 26th international conference on world
  wide web}, pages 173--182, 2017.

\bibitem[Tan et~al.(2014)Tan, Liu, Chen, Xiong, and Wu]{tan2014object}
C.~Tan, Q.~Liu, E.~Chen, H.~Xiong, and X.~Wu.
\newblock Object-oriented travel package recommendation.
\newblock \emph{ACM Transactions on Intelligent Systems and Technology (TIST)},
  5\penalty0 (3):\penalty0 1--26, 2014.

\bibitem[Vandenberghe(2010)]{vandenberghe2010cvxopt}
L.~Vandenberghe.
\newblock The cvxopt linear and quadratic cone program solvers.
\newblock \emph{Online: http://cvxopt. org/documentation/coneprog. pdf}, 2010.

\end{thebibliography}


\begin{thebibliography}{66}
\providecommand{\natexlab}[1]{#1}
\providecommand{\url}[1]{\texttt{#1}}
\providecommand{\urlprefix}{URL }

\bibitem[{Adomavicius \protect\BIBand{} Tuzhilin(2005)}]{adomavicius2005toward}
Adomavicius G, Tuzhilin A (2005) Toward the next generation of recommender
  systems: A survey of the state-of-the-art and possible extensions. \emph{IEEE
  Transactions on Knowledge and Data Engineering} 17(6):734--749.

\bibitem[{Backstrom \protect\BIBand{} Leskovec(2011)}]{backstrom2011supervised}
Backstrom L, Leskovec J (2011) Supervised random walks: predicting and
  recommending links in social networks. \emph{Proceedings of the Fourth ACM
  International Conference on Web Search and Data Mining (WSDM'11)}, 635--644.

\bibitem[{Benchettara et~al.(2010)Benchettara, Kanawati, \protect\BIBand{}
  Rouveirol}]{benchettara2010supervised}
Benchettara N, Kanawati R, Rouveirol C (2010) A supervised machine learning
  link prediction approach for academic collaboration recommendation.
  \emph{Proceedings of the Fourth ACM Conference on Recommender Systems
  (RecSys'10)}, 253--256.

\bibitem[{Benson(2002)}]{benson2002global}
Benson H (2002) Global optimization algorithm for the nonlinear sum of ratios
  problem. \emph{Journal of Optimization Theory and Applications} 112(1):1--29.

\bibitem[{Boim et~al.(2011)Boim, Milo, \protect\BIBand{}
  Novgorodov}]{boim2011diversification}
Boim R, Milo T, Novgorodov S (2011) Diversification and refinement in
  collaborative filtering recommender. \emph{Proceedings of the 20th ACM
  International Conference on Information and Knowledge Management (CIKM'11)},
  739--744.

\bibitem[{Boshkovska et~al.(2015)Boshkovska, Ng, Zlatanov, \protect\BIBand{}
  Schober}]{boshkovska2015practical}
Boshkovska E, Ng DWK, Zlatanov N, Schober R (2015) Practical non-linear energy
  harvesting model and resource allocation for swipt systems. \emph{IEEE
  Communications Letters} 19(12):2082--2085.

\bibitem[{Boyd et~al.(2004)Boyd, Boyd, \protect\BIBand{}
  Vandenberghe}]{boyd2004convex}
Boyd S, Boyd SP, Vandenberghe L (2004) \emph{Convex Optimization} (Cambridge
  university press).

\bibitem[{Burt et~al.(1998)Burt, Jannotta, \protect\BIBand{}
  Mahoney}]{burt1998personality}
Burt RS, Jannotta JE, Mahoney JT (1998) Personality correlates of structural
  holes. \emph{Social Networks} 20(1):63--87.

\bibitem[{Castells et~al.(2015)Castells, Hurley, \protect\BIBand{}
  Vargas}]{castells2015novelty}
Castells P, Hurley NJ, Vargas S (2015) Novelty and diversity in recommender
  systems. \emph{Recommender Systems Handbook}, 881--918.

\bibitem[{Chen et~al.(2010)Chen, Batson, \protect\BIBand{}
  Dang}]{chen2010applied}
Chen DS, Batson RG, Dang Y (2010) \emph{Applied Integer Programming} (Wiley
  Online Library).

\bibitem[{Chen et~al.(2009)Chen, Geyer, Dugan, Muller, \protect\BIBand{}
  Guy}]{chen2009make}
Chen J, Geyer W, Dugan C, Muller M, Guy I (2009) Make new friends, but keep the
  old: recommending people on social networking sites. \emph{Proceedings of the
  SIGCHI Conference on Human Factors in Computing Systems (SIGCHI'09)},
  201--210.

\bibitem[{Chen et~al.(2018)Chen, Zhang, \protect\BIBand{} Zhou}]{chen2018fast}
Chen L, Zhang G, Zhou E (2018) Fast greedy map inference for determinantal
  point process to improve recommendation diversity. \emph{Advances in Neural
  Information Processing Systems (NIPS'18)}, 5622--5633.

\bibitem[{Davenport \protect\BIBand{} Patil(2012)}]{davenport2012data}
Davenport TH, Patil D (2012) Data scientist. \emph{Harvard Business Review}
  90(5):70--76.

\bibitem[{De~Boom et~al.(2015)De~Boom, Van~Canneyt, Bohez, Demeester,
  \protect\BIBand{} Dhoedt}]{de2015learning}
De~Boom C, Van~Canneyt S, Bohez S, Demeester T, Dhoedt B (2015) Learning
  semantic similarity for very short texts. \emph{2015 IEEE International
  Conference on Data Mining Workshop (ICDMW'15)}, 1229--1234.

\bibitem[{Debnath et~al.(2016)Debnath, Tripathi, \protect\BIBand{}
  Elmasri}]{debnath2016preference}
Debnath M, Tripathi PK, Elmasri R (2016) Preference-aware poi recommendation
  with temporal and spatial influence. \emph{The Twenty-Ninth International
  Flairs Conference}.

\bibitem[{Ekstrand et~al.(2014)Ekstrand, Harper, Willemsen, \protect\BIBand{}
  Konstan}]{ekstrand2014user}
Ekstrand MD, Harper FM, Willemsen MC, Konstan JA (2014) User perception of
  differences in recommender algorithms. \emph{Proceedings of the 8th ACM
  Conference on Recommender Systems (RecSys'14)}, 161--168.

\bibitem[{Fang \protect\BIBand{} Hu(2016)}]{fang2016top}
Fang X, Hu PJ (2016) Top persuader prediction for social networks. \emph{MIS
  Quarterly} 42(1):63--82.

\bibitem[{Fang et~al.(2013{\natexlab{a}})Fang, Hu, Li, \protect\BIBand{}
  Tsai}]{fang2013predicting}
Fang X, Hu PJ, Li Z, Tsai W (2013{\natexlab{a}}) Predicting adoption
  probabilities in social networks. \emph{Information Systems Research}
  24(1):128--145.

\bibitem[{Fang et~al.(2013{\natexlab{b}})Fang, Sheng, \protect\BIBand{}
  Goes}]{fang2013right}
Fang X, Sheng ORL, Goes P (2013{\natexlab{b}}) When is the right time to
  refresh knowledge discovered from data? \emph{Operations Research}
  61(1):32--44.

\bibitem[{Gong et~al.(2014)Gong, Talwalkar, Mackey, Huang, Shin, Stefanov, Shi,
  \protect\BIBand{} Song}]{GongSVM}
Gong NZ, Talwalkar A, Mackey L, Huang L, Shin ECR, Stefanov E, Shi ER, Song D
  (2014) Joint link prediction and attribute inference using a social-attribute
  network. \emph{ACM Transactions on Intelligent Systems and Technology} 5(2).

\bibitem[{Gong et~al.(2012)Gong, Xu, Huang, Mittal, Stefanov, Sekar,
  \protect\BIBand{} Song}]{gong2012evolution}
Gong NZ, Xu W, Huang L, Mittal P, Stefanov E, Sekar V, Song D (2012) Evolution
  of social-attribute networks: measurements, modeling, and implications using
  google+. \emph{Proceedings of the 2012 ACM Internet Measurement Conference
  (IMC'12)}, 131--144.

\bibitem[{Gregor \protect\BIBand{} Hevner(2013)}]{gregor2013positioning}
Gregor S, Hevner AR (2013) Positioning and presenting design science research
  for maximum impact. \emph{MIS Quarterly} 37(2):337--355.

\bibitem[{Grover \protect\BIBand{} Leskovec(2016)}]{grover2016node2vec}
Grover A, Leskovec J (2016) node2vec: Scalable feature learning for networks.
  \emph{Proceedings of the 22nd ACM SIGKDD International Conference on
  Knowledge Discovery and Data Mining (KDD'16)}, 855--864.

\bibitem[{Gupta(2018)}]{gupta_traits_2018}
Gupta A (2018) Traits of {Successful} {Research} {Contributions} for
  {Publication} in {ISR}: {Some} {Thoughts} for {Authors} and {Reviewers}.
  \emph{Information Systems Research} 29(4):779--786.

\bibitem[{Hamilton et~al.(2017{\natexlab{a}})Hamilton, Ying, \protect\BIBand{}
  Leskovec}]{hamilton2017inductive}
Hamilton W, Ying Z, Leskovec J (2017{\natexlab{a}}) Inductive representation
  learning on large graphs. \emph{Advances in Neural Information Processing
  Systems (NIPS'17)}, 1024--1034.

\bibitem[{Hamilton et~al.(2017{\natexlab{b}})Hamilton, Ying, \protect\BIBand{}
  Leskovec}]{hamilton2017representation}
Hamilton WL, Ying R, Leskovec J (2017{\natexlab{b}}) Representation learning on
  graphs: Methods and applications. \emph{IEEE Data Engineering Bulletin} .

\bibitem[{He et~al.(2019)He, Fang, Liu, \protect\BIBand{} Li}]{he2019mobile}
He J, Fang X, Liu H, Li X (2019) Mobile app recommendation: an
  involvement-enhanced approach. \emph{MIS Quarterly} 43(3):827--849.

\bibitem[{Kaminskas \protect\BIBand{} Bridge(2016)}]{kaminskas2016diversity}
Kaminskas M, Bridge D (2016) Diversity, serendipity, novelty, and coverage: a
  survey and empirical analysis of beyond-accuracy objectives in recommender
  systems. \emph{ACM Transactions on Interactive Intelligent Systems (TiiS)}
  7(1):1--42.

\bibitem[{Karlof(2005)}]{karlof2005integer}
Karlof JK (2005) \emph{Integer Programming: Theory and Practice} (CRC Press).

\bibitem[{Kempe et~al.(2003)Kempe, Kleinberg, \protect\BIBand{}
  Tardos}]{kempe2003maximizing}
Kempe D, Kleinberg J, Tardos {\'E} (2003) Maximizing the spread of influence
  through a social network. \emph{Proceedings of the Ninth ACM SIGKDD
  International Conference on Knowledge Discovery and Data Mining (KDD'03)},
  137--146.

\bibitem[{Kipf \protect\BIBand{} Welling(2017)}]{kipf2016semi}
Kipf TN, Welling M (2017) Semi-supervised classification with graph
  convolutional networks. \emph{Proceedings of the Fifth International
  Conference on Learning Representation (ICLR'17)} .

\bibitem[{Laakasuo et~al.(2017)Laakasuo, Rotkirch, Berg, \protect\BIBand{}
  Jokela}]{laakasuo2017company}
Laakasuo M, Rotkirch A, Berg V, Jokela M (2017) The company you keep:
  Personality and friendship characteristics. \emph{Social Psychological and
  Personality Science} 8(1):66--73.

\bibitem[{Lakhotia et~al.(2020)Lakhotia, Kannan, Pati, \protect\BIBand{}
  Prasanna}]{lakhotia2020gpop}
Lakhotia K, Kannan R, Pati S, Prasanna V (2020) Gpop: A scalable cache-and
  memory-efficient framework for graph processing over parts. \emph{ACM
  Transactions on Parallel Computing (TOPC)} 7(1):1--24.

\bibitem[{Li et~al.(2017{\natexlab{a}})Li, Fang, Bai, \protect\BIBand{}
  Sheng}]{li2017utility}
Li Z, Fang X, Bai X, Sheng ORL (2017{\natexlab{a}}) Utility-based link
  recommendation for online social networks. \emph{Management Science}
  63(6):1938--1952.

\bibitem[{Li et~al.(2017{\natexlab{b}})Li, Fang, \protect\BIBand{}
  Sheng}]{li2017survey}
Li Z, Fang X, Sheng ORL (2017{\natexlab{b}}) A survey of link recommendation
  for social networks: Methods, theoretical foundations, and future research
  directions. \emph{ACM Transactions on Management Information Systems
  (TMIS'17)} 9(1):1--26.

\bibitem[{Li et~al.(2020)Li, Ge, \protect\BIBand{} Bai}]{li2020will}
Li Z, Ge Y, Bai X (2020) What will be popular next? predicting hotspots in
  two-mode social networks. \emph{MIS Quarterly, forthcoming} .

\bibitem[{Liben-Nowell \protect\BIBand{} Kleinberg(2007)}]{liben2007link}
Liben-Nowell D, Kleinberg J (2007) The link-prediction problem for social
  networks. \emph{Journal of the American Society for Information Science and
  Technology} 58(7):1019--1031.

\bibitem[{Lu et~al.(2016)Lu, Yu, Li, \protect\BIBand{} Wei}]{lu2016exploring}
Lu C, Yu JX, Li RH, Wei H (2016) Exploring hierarchies in online social
  networks. \emph{IEEE Transactions on Knowledge and Data Engineering}
  28(8):2086--2100.

\bibitem[{Mikolov et~al.(2013)Mikolov, Chen, Corrado, \protect\BIBand{}
  Dean}]{mikolov2013efficient}
Mikolov T, Chen K, Corrado G, Dean J (2013) Efficient estimation of word
  representations in vector space. \emph{arXiv preprint arXiv:1301.3781} .

\bibitem[{Padmanabhan et~al.(2022)Padmanabhan, Fang, Sahoo, \protect\BIBand{}
  Burton-Jones}]{padmanabhan2022machine}
Padmanabhan B, Fang X, Sahoo N, Burton-Jones A (2022) Machine learning in
  information systems research. \emph{MIS Quarterly} 46(1):iii--xix.

\bibitem[{Pareja et~al.(2020)Pareja, Domeniconi, Chen, Ma, Suzumura, Kanezashi,
  Kaler, Schardl, \protect\BIBand{} Leiserson}]{pareja2020evolvegcn}
Pareja A, Domeniconi G, Chen J, Ma T, Suzumura T, Kanezashi H, Kaler T, Schardl
  TB, Leiserson CE (2020) Evolvegcn: Evolving graph convolutional networks for
  dynamic graphs. \emph{Proceedings of the Thirth-Fourth AAAI Conference on
  Artificial Intelligence (AAAI'20)}, 5363--5370.

\bibitem[{Perozzi et~al.(2014)Perozzi, Al-Rfou, \protect\BIBand{}
  Skiena}]{perozzi2014deepwalk}
Perozzi B, Al-Rfou R, Skiena S (2014) Deepwalk: Online learning of social
  representations. \emph{Proceedings of the 20th ACM SIGKDD International
  Conference on Knowledge Discovery and Data Mining (KDD'14)}, 701--710.

\bibitem[{Qiu \protect\BIBand{} Kumar(2017)}]{qiu2017understanding}
Qiu L, Kumar S (2017) Understanding voluntary knowledge provision and content
  contribution through a social-media-based prediction market: A field
  experiment. \emph{Information Systems Research} 28(3):529--546.

\bibitem[{Rao(2019)}]{rao2019engineering}
Rao SS (2019) \emph{Engineering Optimization: Theory and Practice} (John Wiley
  \& Sons).

\bibitem[{Rendle et~al.(2009)Rendle, Freudenthaler, Gantner, \protect\BIBand{}
  Schmidt-Thieme}]{rendle2009bpr}
Rendle S, Freudenthaler C, Gantner Z, Schmidt-Thieme L (2009) Bpr: Bayesian
  personalized ranking from implicit feedback. \emph{Proceedings of the
  Twenty-Fifth Conference on Uncertainty in Artificial Intelligence}, 452--461.

\bibitem[{Rivera et~al.(2010)Rivera, Soderstrom, \protect\BIBand{}
  Uzzi}]{rivera2010dynamics}
Rivera MT, Soderstrom SB, Uzzi B (2010) Dynamics of dyads in social networks:
  Assortative, relational, and proximity mechanisms. \emph{Annual Review of
  Sociology} 36:91--115.

\bibitem[{Sanz-Cruzado \protect\BIBand{} Castells(2018)}]{sanz2018enhancing}
Sanz-Cruzado J, Castells P (2018) Enhancing structural diversity in social
  networks by recommending weak ties. \emph{Proceedings of the 12th ACM
  Conference on Recommender Systems (RecSys'18)}, 233--241.

\bibitem[{Schaible \protect\BIBand{} Shi(2003)}]{schaible2003fractional}
Schaible S, Shi J (2003) Fractional programming: the sum-of-ratios case.
  \emph{Optimization Methods and Software} 18(2):219--229.

\bibitem[{Schifanella et~al.(2010)Schifanella, Barrat, Cattuto, Markines,
  \protect\BIBand{} Menczer}]{schifanella2010folks}
Schifanella R, Barrat A, Cattuto C, Markines B, Menczer F (2010) Folks in
  folksonomies: social link prediction from shared metadata. \emph{Proceedings
  of the Third ACM International Conference on Web Search and Data Mining
  (WSDM'10)}, 271--280.

\bibitem[{Shi et~al.(2012)Shi, Zhao, Wang, Larson, \protect\BIBand{}
  Hanjalic}]{shi2012adaptive}
Shi Y, Zhao X, Wang J, Larson M, Hanjalic A (2012) Adaptive diversification of
  recommendation results via latent factor portfolio. \emph{Proceedings of the
  35th international ACM SIGIR conference on Research and development in
  information retrieval}, 175--184.

\bibitem[{Shriver et~al.(2013)Shriver, Nair, \protect\BIBand{}
  Hofstetter}]{shriver2013social}
Shriver SK, Nair HS, Hofstetter R (2013) Social ties and user-generated
  content: Evidence from an online social network. \emph{Management Science}
  59(6):1425--1443.

\bibitem[{Sibley \protect\BIBand{} Duckitt(2008)}]{sibley2008personality}
Sibley CG, Duckitt J (2008) Personality and prejudice: A meta-analysis and
  theoretical review. \emph{Personality and Social Psychology Review}
  12(3):248--279.

\bibitem[{Su et~al.(2013)Su, Yin, Chen, \protect\BIBand{} Yu}]{su2013set}
Su R, Yin L, Chen K, Yu Y (2013) Set-oriented personalized ranking for
  diversified top-n recommendation. \emph{Proceedings of the 7th ACM Conference
  on Recommender Systems}, 415--418.

\bibitem[{Tropp \protect\BIBand{} Bianchi(2006)}]{tropp2006valuing}
Tropp LR, Bianchi RA (2006) Valuing diversity and interest in intergroup
  contact. \emph{Journal of Social Issues} 62(3):533--551.

\bibitem[{Tulin et~al.(2018)Tulin, Lancee, \protect\BIBand{}
  Volker}]{tulin2018personality}
Tulin M, Lancee B, Volker B (2018) Personality and social capital. \emph{Social
  Psychology Quarterly} 81(4):295--318.

\bibitem[{Veli{\v{c}}kovi{\'c} et~al.(2018)Veli{\v{c}}kovi{\'c}, Cucurull,
  Casanova, Romero, Lio, \protect\BIBand{} Bengio}]{velivckovic2017graph}
Veli{\v{c}}kovi{\'c} P, Cucurull G, Casanova A, Romero A, Lio P, Bengio Y
  (2018) Graph attention networks. \emph{Proceedings of the Sixth International
  Conference on Learning Representation (ICLR'18)} .

\bibitem[{Wang et~al.(2011)Wang, Pedreschi, Song, Giannotti, \protect\BIBand{}
  Barabasi}]{wang2011human}
Wang D, Pedreschi D, Song C, Giannotti F, Barabasi AL (2011) Human mobility,
  social ties, and link prediction. \emph{Proceedings of the 17th ACM SIGKDD
  International Conference on Knowledge Discovery and Data Mining (KDD'11)},
  1100--1108.

\bibitem[{Wu et~al.(2019)Wu, Liu, Miao, Zhao, Guan, \protect\BIBand{}
  Tang}]{wu2019recent}
Wu Q, Liu Y, Miao C, Zhao Y, Guan L, Tang H (2019) Recent advances in
  diversified recommendation. \emph{arXiv preprint arXiv:1905.06589} .

\bibitem[{Wu et~al.(2020)Wu, Pan, Chen, Long, Zhang, \protect\BIBand{}
  Philip}]{wu2020comprehensive}
Wu Z, Pan S, Chen F, Long G, Zhang C, Philip SY (2020) A comprehensive survey
  on graph neural networks. \emph{IEEE Transactions on Neural Networks and
  Learning Systems} .

\bibitem[{Zangerle et~al.(2020)Zangerle, Pichl, \protect\BIBand{}
  Schedl}]{zangerle2020user}
Zangerle E, Pichl M, Schedl M (2020) User models for culture-aware music
  recommendation: fusing acoustic and cultural cues. \emph{Transactions of the
  International Society for Music Information Retrieval} 3(1).

\bibitem[{Zeng \protect\BIBand{} Xie(2008)}]{zeng2008preference}
Zeng Z, Xie Y (2008) A preference-opportunity-choice framework with
  applications to intergroup friendship. \emph{American Journal of Sociology}
  114(3):615--648.

\bibitem[{Zhang \protect\BIBand{} Chen(2018)}]{zhang2018link}
Zhang M, Chen Y (2018) Link prediction based on graph neural networks.
  \emph{Advances in Neural Information Processing Systems (NIPS'18)},
  5165--5175.

\bibitem[{Zhang \protect\BIBand{} Hurley(2008)}]{zhang2008avoiding}
Zhang M, Hurley N (2008) Avoiding monotony: improving the diversity of
  recommendation lists. \emph{Proceedings of the 2008 ACM Conference on
  Recommender Systems (RecSys'08)}, 123--130.

\bibitem[{Zhang \protect\BIBand{} Hurley(2009)}]{zhang2009novel}
Zhang M, Hurley N (2009) Novel item recommendation by user profile
  partitioning. \emph{2009 IEEE/WIC/ACM International Joint Conference on Web
  Intelligence and Intelligent Agent Technology (WI-IAT'09)}, volume~1,
  508--515.

\bibitem[{Zheleva et~al.(2008)Zheleva, Getoor, Golbeck, \protect\BIBand{}
  Kuter}]{zheleva2008using}
Zheleva E, Getoor L, Golbeck J, Kuter U (2008) Using friendship ties and family
  circles for link prediction. \emph{International Workshop on Social Network
  Mining and Analysis}, 97--113.

\bibitem[{Ziegler et~al.(2005)Ziegler, McNee, Konstan, \protect\BIBand{}
  Lausen}]{ziegler2005improving}
Ziegler CN, McNee SM, Konstan JA, Lausen G (2005) Improving recommendation
  lists through topic diversification. \emph{Proceedings of the 14th
  International Conference on World Wide Web (WWW'05)}, 22--32.

\end{thebibliography}

%
\newpage

\pagenumbering{arabic}
\renewcommand*{\thepage}{ec\arabic{page}}
\renewcommand{\theequation}{A.\arabic{equation}}
\setcounter{equation}{0}

\appendix
\begin{bibunit}[abbrvnat]

{\noindent \Large \textbf{Appendix A. Proof of Theorem 1}}
\vspace{0.4cm}

\noindent The DPA-LR problem is formulated as (A.1):
\begin{equation}
\begin{aligned}
& \underset{\vectory}{\text{maximize}}
& & \sum_{h=1}^{H}  \frac{ \vectord^{h^T} \matrixC^h\vectory}{\| \vectord^h \| \|\matrixC^h\vectory\|} \\
& \text{subject to}
& & \textbf{1}^T\vectory=k   \\
&&& y_j \in \{0, 1\},\ j=1,2,\cdots,m
\end{aligned}
\end{equation}

\vspace{-2mm}

\noindent where $\matrixC^h\vectory=\vectorr^h$ and $\matrixC^h$ denotes a user's candidate profile matrix in dimension $h$, $h=1,2, \dots, H.$  
It suffices to show that the DPA-LR problem is NP-hard if we can prove that the problem is NP-hard for $H=1$. Next, we focus on showing the hardness of the problem for $H=1$.
Given $H=1$, we can drop the superscript $h$ from the problem formulation. Let $\matrixQ=\matrixC^T\matrixC$ and denote $\bar{\vectorp}^T=\frac{\vectord^T\matrixC}{\|\vectord\|}$. We can rewrite the DPA-LR problem for $H=1$ as:
\begin{equation}
\begin{aligned}
& \underset{\vectory}{\text{maximize}}
& & \frac{ \bar{\vectorp}^T \vectory}{\sqrt{\vectory^T\matrixQ\vectory}} \\
& \text{subject to}
& & \textbf{1}^T\vectory=k   \\
&&& y_j \in \{0, 1\},\ j=1,2,\cdots,m
\end{aligned}
\end{equation}

Consider the k-clique problem, a well known NP-complete problem. Given an undirected graph $\mathcal{G} = (\mathbb{V}, \mathbb{E})$, where $\mathbb{V}$ and $\mathbb{E}$ represent the sets of nodes and edges, respectively.
A k-clique is a fully connected sub-graph with exact $k$ vertices. The k-clique problem decides if $\mathcal{G}$ contains a k-clique as one of its sub-graphs.
To prove problem (A.2) is NP-hard, we show that the k-clique problem is a special case of the problem. Specifically, for problem (A.2), let $m=|\mathbb{V}|$, $\bar{p}_i=1$, $i=1,\cdots, m$, $\matrixQ = (q_{ij})$, and

\vspace{-1mm}
\begin{equation}
q_{ij} = \begin{cases}
1 & \quad \text{if edge} (v_i, \ v_j) \in \mathbb{E},  \\
2 & \quad \text{otherwise},
\end{cases}
\end{equation}

\noindent where $\bar{p}_i$ is an element of vector $\bar{\vectorp}$.
Denote $\vectory^*$ as an optimal solution of problem (A.2). 
It is easy to verify that $\mathcal{G}$ contains a k-clique if and only if 
$$
\frac{\bar{\vectorp}^T\vectory^*}{\sqrt{\vectory^{*T}\matrixQ\vectory^*}} =
\frac{k}{\sqrt{k(k+1)}}
$$ 
and the vertices of the k-clique are those with $y^*_i= 1$.


{\noindent \Large \textbf{Appendix B. Proof of Theorem 2}}
\vspace{0.4cm}

\noindent To prove Theorem 2, we compare the KKT conditions for problem (3) and those for problem (4).  
Let $\nu$, $\vectormu \in \mathbb{R}^{2m}$, $\vectormu \succeq \boldsymbol{0}$, and $\gamma_h$, $h=1,2, \cdots, H$ be the dual 
variables associated with the respective constraints in problem (3). 
Then the Lagrangian function of the problem is given by:
\vspace{-2mm}
\begin{equation}
\mathcal{L}(\vectory,\ \vectorbeta,\ \vectorgamma,\ \vectormu,\ \nu) =  -\boldsymbol{1}^T\vectorbeta + \nu(\boldsymbol{1}^T \vectory-k) + \vectormu^T(\matrixA \vectory-\vectorb) + \sum_{h=1}^{H} \gamma_h(\beta_h\|\matrixC^h\vectory\| - \bar{\vectord}^{h^T}\matrixC^h\vectory).
\end{equation}

\vspace{-3mm}

\noindent Derived from the Lagrangian function, the KKT conditions for problem (3) can be written as:
\begin{equation}
\frac{\partial \mathcal{L}}{\partial y_j} = \sum_{h=1}^{H}\gamma_h(\frac{\beta_h(\matrixC^{h^T}\matrixC^h)_{j, :}\vectory}{\|\matrixC^h\vectory\|} - \bar{\vectord}^{h^T}\matrixC^h_{:,j}) + (\mu_j+\mu_{m+j}) +\nu = 0, \ \ j=1,2, \cdots,m
\end{equation}
\vspace{-5mm}
\begin{equation}
\frac{\partial \mathcal{L}}{\partial \beta_h} = -1+\gamma_h \|\matrixC^h \vectory \|=0, \ \ h=1,2, \cdots, H
\end{equation}
\vspace{-5mm}
\begin{equation}
\gamma_h(\beta_h\|\matrixC^h\vectory\|-\bar{\vectord}^{h^T}\matrixC^h\vectory) = 0, \ \ h=1,2, \cdots, H
\end{equation}
\vspace{-5mm}
\begin{equation}
\vectormu^{T}(\matrixA\vectory-\vectorb)=\boldsymbol{0}
\end{equation}
\vspace{-5mm}
\begin{equation}
\matrixA\vectory-\vectorb\preceq \boldsymbol{0}
\end{equation}
\vspace{-5mm}
\begin{equation}
\boldsymbol{1}^T\vectory-k=0
\end{equation}
\vspace{-5mm}
\begin{equation}
\mu_j\geq0, \  j=1,2, \cdots, 2m
\end{equation}
In these conditions, notation $M_{j,:}$ denotes the the $j$th row of matrix $M$, and notation $M_{:,j}$ denotes the the $j$th column of matrix $M$. According to \citep{boyd2004convex}, $(\vectory,\vectorbeta)$ is a stationary point solution to  problem (3) if and only if it satisfies the KKT conditions (A.5)-(A.11). 
Notice that condition (A.6) is equivalent to condition (6) in Theorem 2. In addition, it implies that $\gamma_h >0$ for all $h=1, \ldots, H$. It follows that condition (A.7) is equivalent to condition (5) in Theorem 2.

Following the same process, we can also write out the KKT conditions for problem (4) by introducing dual variables $\nu$ and $\vectormu \in \mathbb{R}^{2m}$, $\vectormu \succeq \boldsymbol{0}$, for its two sets of constraints.
It turns out that the KKT conditions for problem (4) are (A.5) and (A.8)-(A.11). Only conditions (A.6) and (A.7) are missing in comparison to the KKT conditions for problem (3). Note that 
problem (4) is a convex optimization problem and the KKT conditions are sufficient and necessary to the optimal solution \citep{boyd2004convex}. Theorem 2 then follows immediately.

\newpage

\setcounter{table}{0} \renewcommand{\thetable}{A.\arabic{table}}

{\noindent \Large \textbf{Appendix C. Time Complexity Analysis of DPA-LR}}
\vspace{0.4cm}

\noindent We analyze the time complexity of our proposed method (DPA-LR).
Table A.1 summarizes the time complexity of each computation step of our method. In this table, $m$ is the number of a user's candidate friends, $n_h$ is the number of unique values for profile dimension $h$, and $H$ is the number of profile dimensions. For example, in the Google+ data set, there are four different profile dimensions (i.e., major, school, employer, place lived) and $H=4$.     

\begin{table}[H]
{\small
	\begin{center}
		
		\captionsetup{justification=centering,margin=1cm}
		\caption{\textbf{Time Complexity Analysis of DPA-LR by Computation Steps}}
		\label{tab:table 1}
		\begin{tabular}{l|c} 
			\hline
			
			Computation Step & Time Complexity\\
			\hline
			$l=1$.  \hspace{0.5cm}   //$l$: iteration count & $\mathcal{O}(1)$ \\
			Randomly initialize parameters $(\vectorgamma^l, \ \vectorbeta^l)$.& $\mathcal{O}(2H)$ \\
			Compute $\vectory^l$ by solving problem (4) with $(\vectorgamma^l, \ \vectorbeta^l)$.& $\mathcal{O}(m^{\frac{5}{2}}\log\frac{m}{\eta})$\\
			Compute errors $\vectordelta^l$ according to Equations (7) and (8).& $\mathcal{O}(m\sum_{h=1}^{H}n_h)$\\
			\textbf{While} Not($ \|\vectordelta^l\|< \varepsilon$)& \\ 
			$\hspace{0.6cm}$ Obtain parameters $(\vectorgamma^{l+1}, \ \vectorbeta^{l+1})$ according to Equations (9) and (10).& $\mathcal{O}(m\sum_{h=1}^{H}n_h)$ \\
			$\hspace{0.75cm}$Compute $\vectory^{l+1}$ by solving problem (4) with $(\vectorgamma^{l+1}, \ \vectorbeta^{l+1})$.& $\mathcal{O}(m^{\frac{5}{2}}\log\frac{m}{\eta})$\\
			$\hspace{0.75cm}$Compute errors $\vectordelta^{l+1}$ according to Equations (7) and (8).& $\mathcal{O}(m\sum_{h=1}^{H}n_h)$\\
			$\hspace{0.75cm}$ $l=l+1$.& $\mathcal{O}(1)$\\
			\textbf{End While}&\\
			$\vectory^s=\vectory^{l}$. & $\mathcal{O}(m)$\\
			Sort entries in $\vectory^s$ and set the top-k entries to 1 and the rest to 0. & $\mathcal{O}(m\log m)$\\		    
			\hline
		\end{tabular}
	\end{center}}
\end{table}
\vspace{-0.4cm}

In particular, we detail the time complexity of the third computation step in the table, i.e., solving problem (4), which is a key computation step of our method. To solve problem (4), we employ the python package CVXOPT\footnote{\url{https://cvxopt.org/userguide/intro.html}}, which transforms problem (4) into its equivalent Second Order Cone Program (SOCP) problem: 
\begin{equation*}
	\begin{aligned}
		& \underset{\vectory}{\text{maximize}}
		& & \textbf{1}^T\vectorbeta + \sum_{h=1}^{H}\gamma_h\bar{\vectord}^{h^T}\matrixC^h\vectory + \sum_{h=1}^{H}\mathcal{S}_h \\
		& \text{subject to}
		& & \textbf{1}^T\vectory-k=0 \\
		&&& \matrixA\vectory-\vectorb \preceq \textbf{0} \\
		&&& \|\matrixC^h\vectory\| \leq -\frac{\mathcal{S}_h}{\gamma_h\beta_h}, \ h=1,2,\cdots, H.
	\end{aligned}
\end{equation*}

CVXOPT solves the above SOCP problem using the interior point method \citep{vandenberghe2010cvxopt} with a time complexity of $\mathcal{O}(m^2\sqrt{\nu}\log\frac{\nu}{\eta})$ \citep{griva2009linear, bubeck2014convex}, where $\nu$ is the number of inequality constraints in the SOCP problem and $\eta$ denotes the given error tolerance. In particular, $\nu=2m+H$ because $\matrixA\vectory-\vectorb \preceq \textbf{0}$ is a set of $2m$ inequalities and there is an inequality constraint $\|\matrixC^h\vectory\| \leq -\frac{\mathcal{S}_h}{\gamma_h\beta_h}$ for each profile dimension $h=1,2,\cdots, H$. By substituting $\nu$ with $2m+H$, keeping dominant terms, and dropping constants, the time complexity of solving problem (4) is $\mathcal{O}(m^{\frac{5}{2}}\log\frac{m}{\eta})$.

By summing up the time complexity of each computation step of our method listed in Table A.1, keeping dominant terms, and dropping constants, the time complexity of our method is 
\[ \mathcal{O}(L(m\sum_{h=1}^{H}n_h+m^{5/2}\log\frac{m}{\eta})), \]
where $L$ denotes the number of iterations needed for our method to converge. Using the Google+ data set, we empirically show that, on average, our method converges after $7$ iterations, i.e., $L=7$ (see Section 5.4).

\newpage

{\noindent \Large \textbf{Appendix D. Empirical Evaluation with Another Online Social \\}}

\vspace{-3mm}

{\noindent \Large \textbf{Network Data Set}}

\vspace{0.4cm}

\noindent We evaluated the performance of the methods listed in Table 4 using another large-scale data set collected from a major U.S. online social network. The data set contains profiles of the network users as well as their linkages in three consecutive months during the early stage of this social network, namely time periods 0, 1, and 2. Table A.2 summarizes the linkage data. As reported in the table, there are 824,178 users connected by 3,503,082 undirected links in time period 0. Due to privacy concerns, a user is described by a set of encoded terms, each of which represents a characteristic of the user. There are 3,370 unique terms across the users. Hence, we represented a user's profile as a vector of size 3,370, an element of which was set to 1 if its corresponding term appeared in the user's profile or set to 0 otherwise. A user's profile contains as many as 477 terms and an average of 5.4 terms.


\begin{table}[H]
	\begin{center}
		\footnotesize{
			\captionsetup{justification=centering,margin=0.5cm}
			\caption{\textbf{Summary Statistics of the Linkage Data}}
			\label{tab:table 1}
			\renewcommand{\arraystretch}{0.8}
			\begin{tabular}{c  @{\hskip 0.15in} c  @{\hskip 0.15in} c} 
				\hline
				\textbf{Time Period} & \textbf{Number of Users} & \textbf{Number of Undirected Links} \\
				\hline
				0 & 824,178 & 3,503,082 \\
				1 & 954,498 & 4,114,564 \\
				2 & 1,095,832 & 4,805,382 \\
				\hline
			\end{tabular}
			\renewcommand{\arraystretch}{0.8}}
	\end{center}
\end{table}
\vspace{-0.5cm}


Following the evaluation procedure in Section 5.2, we compared the performance of the methods by initially setting the number of recommended friends $k=10$ and the diversity weight $\theta$ for methods MMR, MSD, and DPP to 0.5. Table A.3 reports the performance of each method for the four evaluation metrics, with the percentage improvement of our method over a benchmark listed in parentheses under the benchmark's performance.
As shown, our method performs the best in all evaluation metrics. In particular, it outperforms the respective best performing benchmark by 13.70\% in DPMS, 18.01\% in precision, 14.30\% in recall, and 15.96\% in F1.
Applying the paired t-test to the performance results of recommendations by each method shows that our method significantly outperforms each benchmark method across all four metrics ($p<0.001$). 

\begin{table}[H]
	
	\begin{center}
		\footnotesize{
			\captionsetup{justification=centering,margin=0.5cm}
			\caption{\textbf{Performance Comparison between DPA-LR and Benchmarks ($k$=10)}}
			\label{tab:table 1}
			\renewcommand{\arraystretch}{0.8}
			\begin{tabular}{c@{\hskip 0.2in}c@{\hskip 0.2in}c@{\hskip 0.2in}c@{\hskip 0.2in}c} 
				\hline
				\textbf{Methods}  &  \textbf{DPMS} & \textbf{Precision} &  \textbf{Recall} &  \textbf{F1 Score}   \\
				\hline

				DPA-LR 				  & 0.5622  & 0.0375    & 0.1911  & 0.0552  \\
				MMR ($\theta = 0.5$)  & 0.3634  & 0.0208    & 0.1044  & 0.0306  \\[-0.4em]
									  & (54.71\%) & (79.89\%)   & (83.07\%) & (80.59\%) \\
				MSD ($\theta = 0.5$)  & 0.4684  & 0.0299    & 0.1432  & 0.0426  \\[-0.4em]
									  & (20.02\%) & (25.59\%)   & (33.42\%) & (29.71\%) \\
				DPP ($\theta = 0.5$)  & 0.4945  & 0.0309    & 0.1630  & 0.0460  \\[-0.4em]
									  & (13.70\%) & (21.25\%)   & (17.26\%) & (20.12\%) \\
				DiRec 				  & 0.4906  & 0.0318    & 0.1672  & 0.0476  \\[-0.4em]
									  & (14.60\%) & (18.01\%)   & (14.30\%) & (15.96\%)\\

				\hline
			\end{tabular}
			\renewcommand{\arraystretch}{0.8}}
	\end{center}
\vspace{-3mm}
\centering{\footnotesize Note: Percentage improvement of our method over a benchmark listed in parentheses.}
\end{table}

\vspace{-2mm}

We conducted additional experiments for robustness checks by varying $k$ between 6 and 14 and changing the diversity weight $\theta$ for benchmark methods MMR, MSD, and DPP from 0.1 to 0.9. As reported in Tables A.4 and A.5, DPA-LR substantially and significantly outperforms each benchmark in all four metrics, across the investigated values of $k$ and $\theta$ ($p<0.001$). 


\begin{table}[H]
	
	\begin{center}
		\footnotesize{
			\captionsetup{justification=centering,margin=0.5cm}
			\caption{\textbf{Performance Comparison between DPA-LR and Benchmarks: $k$=6 to $k$=14}}
			\label{tab:table 1}
			\renewcommand{\arraystretch}{0.8}
			\begin{tabular}{c@{\hskip 0.2in} c @{\hskip 0.2in}  c@{\hskip 0.2in} c@{\hskip 0.2in} c @{\hskip 0.2in}c@{\hskip 0.2in} c} 
				\hline
				\textbf{Metric} &\textbf{Methods}&\textbf{$k$=6}&\textbf{$k$=8}&\textbf{$k$=10}&\textbf{$k$=12}&\textbf{$k$=14}   \\
				\hline
				DPMS        & DPA-LR & 0.5293  & 0.5486  & 0.5622  & 0.5723  & 0.5801  \\[-0.2em]
				& MMR ($\theta=0.5$)  & 0.3092  & 0.3373  & 0.3634  & 0.3875  & 0.4096  \\[-0.4em]
				&       & (71.18\%) & (62.63\%) & (54.71\%) & (47.70\%) & (41.63\%) \\[-0.2em]
				& MSD ($\theta=0.5$)  & 0.4053  & 0.4399  & 0.4684  & 0.4911  & 0.5094  \\[-0.4em]
				&       & (30.61\%) & (24.71\%) & (20.02\%) & (16.55\%) & (13.89\%) \\[-0.2em]
				& DPP ($\theta=0.5$)  & 0.4448  & 0.4739  & 0.4945  & 0.5102  & 0.5228  \\[-0.4em]
				&       & (19.00\%) & (15.77\%) & (13.70\%) & (12.18\%) & (10.96\%) \\[-0.2em]
				& DiRec & 0.4378  & 0.4688  & 0.4906  & 0.5071  & 0.5199  \\[-0.4em]
				&       & (20.90\%) & (17.03\%) & (14.60\%) & (12.88\%) & (11.58\%) \\
				
				\hline
				Precision   & DPA-LR & 0.0402  & 0.0388  & 0.0375  & 0.0365  & 0.0357  \\[-0.2em]
				& MMR ($\theta=0.5$)   & 0.0226  & 0.0214  & 0.0208  & 0.0207  & 0.0207  \\[-0.4em]
				&       & (77.56\%) & (81.15\%) & (79.89\%) & (76.19\%) & (72.03\%) \\[-0.2em]
				& MSD ($\theta=0.5$)  & 0.0303  & 0.0301  & 0.0299  & 0.0296  & 0.0294  \\[-0.4em]
				&       & (32.49\%) & (28.78\%) & (25.59\%) & (23.49\%) & (21.52\%) \\[-0.2em]
				& DPP ($\theta=0.5$)  & 0.0328  & 0.0318  & 0.0309  & 0.0304  & 0.0299  \\[-0.4em]
				&       & (22.39\%) & (22.07\%) & (21.25\%) & (20.17\%) & (19.15\%) \\[-0.2em]
				& DiRec & 0.0341  & 0.0330  & 0.0318  & 0.0305  & 0.0295  \\[-0.4em]
				&       & (17.69\%) & (17.60\%) & (18.01\%) & (19.50\%) & (20.97\%) \\
				
				\hline
				Recall      & DPA-LR & 0.1250  & 0.1591  & 0.1911  & 0.2212  & 0.2496  \\[-0.2em]
				& MMR ($\theta=0.5$)  & 0.0688  & 0.0859  & 0.1044  & 0.1239  & 0.1439  \\[-0.4em]
				&       & (81.63\%) & (85.24\%) & (83.07\%) & (78.53\%) & (73.46\%) \\[-0.2em]
				& MSD ($\theta=0.5$)  & 0.0877  & 0.1160  & 0.1432  & 0.1692  & 0.1952  \\[-0.4em]
				&       & (42.57\%) & (37.17\%) & (33.42\%) & (30.68\%) & (27.90\%) \\[-0.2em]
				& DPP ($\theta=0.5$)  & 0.1053  & 0.1351  & 0.1630  & 0.1904  & 0.2172  \\[-0.4em]
				&       & (18.77\%) & (17.79\%) & (17.26\%) & (16.13\%) & (14.94\%) \\[-0.2em]
				& DiRec & 0.1107  & 0.1401  & 0.1672  & 0.1912  & 0.2137  \\[-0.4em]
				&       & (12.89\%) & (13.55\%) & (14.30\%) & (15.68\%) & (16.78\%) \\
				
				\hline
				F1 Score    & DPA-LR & 0.0527  & 0.0545  & 0.0552  & 0.0556  & 0.0557  \\[-0.2em]
				& MMR ($\theta=0.5$)  & 0.0295  & 0.0299  & 0.0306  & 0.0315  & 0.0324  \\[-0.4em]
				&       & (78.91\%) & (82.40\%) & (80.59\%) & (76.55\%) & (72.09\%) \\[-0.2em]
				& MSD ($\theta=0.5$)  & 0.0381  & 0.0408  & 0.0426  & 0.0437  & 0.0447  \\[-0.4em]
				&       & (38.49\%) & (33.60\%) & (29.71\%) & (27.11\%) & (24.72\%) \\[-0.2em]
				& DPP ($\theta=0.5$)  & 0.0436  & 0.0452  & 0.0460  & 0.0466  & 0.0471  \\[-0.4em]
				&       & (20.90\%) & (20.64\%) & (20.12\%) & (19.22\%) & (18.31\%) \\[-0.2em]
				& DiRec & 0.0458  & 0.0472  & 0.0476  & 0.0473  & 0.0467  \\[-0.4em]
				&       & (14.99\%) & (15.31\%) & (15.96\%) & (17.59\%) & (19.22\%) \\
				
				\hline
			\end{tabular}
			\renewcommand{\arraystretch}{0.8}}
	\end{center}
\vspace{-3mm}
\centering{\footnotesize Note: Percentage improvement of our method over a benchmark listed in parentheses.}
\end{table}


\begin{table}[H]
	
	\begin{center}
		\footnotesize{
			\captionsetup{justification=centering,margin=0.5cm}
			\caption{\textbf{Performance Comparison between DPA-LR and Benchmarks ($k$=10): $\theta=0.1$ to $\theta=0.9$}}
			\label{tab:table 1}
			\renewcommand{\arraystretch}{0.8}
			\begin{tabular}{c@{\hskip 0.3in}c@{\hskip 0.3in} c @{\hskip 0.3in}c @{\hskip 0.3in}c @{\hskip 0.3in}c } 
				\hline
				\textbf{Method} &  \textbf{Diversity Weight} & \textbf{DPMS} & \textbf{Precision} & \textbf{Recall} & \textbf{F1 Score}   \\
				\hline
				DPA-LR   &		-	   & 0.5622  & 0.0375   & 0.1911   & 0.0552   \\
				
				\hline
				MMR     & $\theta=0.1$ & 0.4467  & 0.0318   & 0.1616   & 0.0473   \\[-0.3em]
				&              & (25.86\%) & (17.88\%)  & (18.28\%)  & (16.69\%)  \\
				& $\theta=0.2$ & 0.4208  & 0.0290   & 0.1473   & 0.0431   \\[-0.3em]
				&              & (33.60\%) & (29.07\%)  & (29.77\%)  & (28.01\%)  \\
				& $\theta=0.3$ & 0.3971  & 0.0259   & 0.1306   & 0.0383   \\[-0.3em]
				&              & (41.60\%) & (44.62\%)  & (46.28\%)  & (44.08\%)  \\
				& $\theta=0.4$ & 0.3785  & 0.0233   & 0.1173   & 0.0343   \\[-0.3em]
				&              & (48.55\%) & (60.89\%)  & (62.92\%)  & (60.86\%)  \\
				& $\theta=0.5$ & 0.3634  & 0.0208   & 0.1044   & 0.0306   \\[-0.3em]
				&              & (54.71\%) & (79.89\%)  & (83.07\%)  & (80.59\%)  \\
				& $\theta=0.6$ & 0.3507  & 0.0192   & 0.0956   & 0.0281   \\[-0.3em]
				&              & (60.31\%) & (94.88\%)  & (99.89\%)  & (96.40\%)  \\
				& $\theta=0.7$ & 0.3389  & 0.0178   & 0.0875   & 0.0259   \\[-0.3em]
				&              & (65.92\%) & (110.37\%) & (118.43\%) & (113.11\%) \\
				& $\theta=0.8$ & 0.3275  & 0.0165   & 0.0800   & 0.0239   \\[-0.3em]
				&              & (71.66\%) & (127.18\%) & (138.97\%) & (131.40\%) \\
				& $\theta=0.9$ & 0.3164  & 0.0157   & 0.0745   & 0.0224   \\[-0.3em]
				&              & (77.70\%) & (139.12\%) & (156.60\%) & (146.55\%) \\
				
				\hline
				MSD     & $\theta=0.1$ & 0.4747  & 0.0306   & 0.1448   & 0.0438   \\[-0.3em]
				&              & (18.45\%) & (22.39\%)  & (31.98\%)  & (26.04\%)  \\
				& $\theta=0.2$ & 0.4798  & 0.0305   & 0.1448   & 0.0434   \\[-0.3em]
				&              & (17.19\%) & (22.82\%)  & (31.94\%)  & (27.19\%)  \\
				& $\theta=0.3$ & 0.4789  & 0.0305   & 0.1447   & 0.0434   \\[-0.3em]
				&              & (17.39\%) & (22.90\%)  & (32.03\%)  & (27.33\%)  \\
				& $\theta=0.4$ & 0.4746  & 0.0302   & 0.1443   & 0.0430   \\[-0.3em]
				&              & (18.47\%) & (24.25\%)  & (32.41\%)  & (28.39\%)  \\
				& $\theta=0.5$ & 0.4684  & 0.0299   & 0.1432   & 0.0426   \\[-0.3em]
				&              & (20.02\%) & (25.59\%)  & (33.42\%)  & (29.71\%)  \\
				& $\theta=0.6$ & 0.4618  & 0.0298   & 0.1410   & 0.0422   \\[-0.3em]
				&              & (21.74\%) & (25.84\%)  & (35.52\%)  & (30.89\%)  \\
				& $\theta=0.7$ & 0.4577  & 0.0296   & 0.1407   & 0.0420   \\[-0.3em]
				&              & (22.84\%) & (26.87\%)  & (35.85\%)  & (31.51\%)  \\
				& $\theta=0.8$ & 0.4563  & 0.0293   & 0.1388   & 0.0415   \\[-0.3em]
				&              & (23.23\%) & (27.90\%)  & (37.65\%)  & (32.98\%)  \\
				& $\theta=0.9$ & 0.4561  & 0.0292   & 0.1383   & 0.0413   \\[-0.3em]
				&              & (23.27\%) & (28.41\%)  & (38.23\%)  & (33.63\%)  \\
				
				\hline
				DPP     & $\theta=0.1$ & 0.4965  & 0.0313   & 0.1637   & 0.0465   \\[-0.3em]
				&              & (13.25\%) & (19.73\%)  & (16.75\%)  & (18.63\%)  \\
				& $\theta=0.2$ & 0.4960  & 0.0312   & 0.1635   & 0.0464   \\[-0.3em]
				&              & (13.36\%) & (20.11\%)  & (16.89\%)  & (19.01\%)  \\
				& $\theta=0.3$ & 0.4955  & 0.0311   & 0.1633   & 0.0462   \\[-0.3em]
				&              & (13.47\%) & (20.48\%)  & (17.00\%)  & (19.36\%)  \\
				& $\theta=0.4$ & 0.4950  & 0.0310   & 0.1631   & 0.0461   \\[-0.3em]
				&              & (13.59\%) & (20.86\%)  & (17.14\%)  & (19.74\%)  \\
				& $\theta=0.5$ & 0.4945  & 0.0309   & 0.1630   & 0.0460   \\[-0.3em]
				&              & (13.70\%) & (21.25\%)  & (17.26\%)  & (20.12\%)  \\
				& $\theta=0.6$ & 0.4925  & 0.0308   & 0.1629   & 0.0458   \\[-0.3em]
				&              & (14.16\%) & (21.62\%)  & (17.35\%)  & (20.48\%)  \\
				& $\theta=0.7$ & 0.4905  & 0.0307   & 0.1626   & 0.0457   \\[-0.3em]
				&              & (14.62\%) & (22.04\%)  & (17.50\%)  & (20.89\%)  \\
				& $\theta=0.8$ & 0.4886  & 0.0306   & 0.1624   & 0.0455   \\[-0.3em]
				&              & (15.08\%) & (22.55\%)  & (17.70\%)  & (21.38\%)  \\
				& $\theta=0.9$ & 0.4866  & 0.0305   & 0.1619   & 0.0453   \\[-0.3em]
				&              & (15.54\%) & (23.12\%)  & (18.01\%)  & (21.91\%) \\
				
				\hline
			\end{tabular}
			\renewcommand{\arraystretch}{0.8}}
	\end{center}
\vspace{-3mm}
\centering{\footnotesize Note: Percentage improvement of our method over a benchmark listed in parentheses.}
\end{table}

\newpage

{\noindent \Large \textbf{Appendix E. Performance Comparison between DPA-LR and  \\}}

\vspace{-3mm}

{\noindent \Large \textbf{State-of-the-art Link Recommendation Methods}}

\vspace{0.4cm}

\noindent We compared the performance between state-of-the-art link recommendation methods and our DPA-LR method by varying the number of recommended friends $k$ between 6 and 14. As reported in the Table A.6 below, DPA-LR substantially and significantly outperforms each of the compared methods in all performance metrics across $k$ ($p<0.001$). 

\begin{table}[H]
	
	\begin{center}
		\small{
			\captionsetup{justification=centering,margin=0.5cm}
			\caption{\textbf{Comparison between DPA-LR and State-of-the-art Link Recommendation Methods: $k$=6 to $k$=14}}
			\label{tab:table 1}
			\renewcommand{\arraystretch}{0.8}
			\begin{tabular}{c@{\hskip 0.2in} c @{\hskip 0.2in}  c@{\hskip 0.2in} c@{\hskip 0.2in} c @{\hskip 0.2in}c@{\hskip 0.2in} c} 
				\hline
				\textbf{Metric} &\textbf{Methods}&\textbf{$k$=6}&\textbf{$k$=8}&\textbf{$k$=10}&\textbf{$k$=12}&\textbf{$k$=14}   \\
				\hline
				DPMS & DPA-LR        & 0.3998   & 0.4100   & 0.4559   & 0.4595   & 0.4612   \\
				& GCN-LR       & 0.1993   & 0.2210   & 0.2389   & 0.2540   & 0.2663   \\[-0.4em]
				&              & (100.59\%) & (85.52\%)  & (90.89\%)  & (80.91\%)  & (73.19\%)  \\
				& GraphSage-LR & 0.1577   & 0.1751   & 0.1895   & 0.2018   & 0.2124   \\[-0.4em]
				&              & (153.58\%) & (134.16\%) & (140.59\%) & (127.69\%) & (117.14\%) \\
				& GAT-LR       & 0.1642   & 0.1833   & 0.1992   & 0.2126   & 0.2243   \\[-0.4em]
				&              & (143.53\%) & (123.68\%) & (128.89\%) & (116.10\%) & (105.59\%) \\
				\hline
				Precision & DPA-LR        & 0.1619  & 0.1577  & 0.1541  & 0.1509  & 0.1479  \\
				& GCN-LR       & 0.1455  & 0.1411  & 0.1380  & 0.1354  & 0.1329  \\[-0.4em]
				&              & (11.25\%) & (11.74\%) & (11.72\%) & (11.43\%) & (11.30\%) \\
				& GraphSage-LR & 0.1271  & 0.1252  & 0.1234  & 0.1216  & 0.1195  \\[-0.4em]
				&              & (27.39\%) & (25.98\%) & (24.94\%) & (24.04\%) & (23.75\%) \\
				& GAT-LR       & 0.1273  & 0.1235  & 0.1206  & 0.1183  & 0.1163  \\[-0.4em]
				&              & (27.13\%) & (27.68\%) & (27.79\%) & (27.49\%) & (27.15\%) \\
				\hline
				Recall    & DPA-LR        & 0.1024  & 0.1302  & 0.1559  & 0.1792  & 0.2015  \\
				& GCN-LR       & 0.0734  & 0.0939  & 0.1147  & 0.1353  & 0.1561  \\[-0.4em]
				&              & (39.41\%) & (38.66\%) & (35.98\%) & (32.51\%) & (29.13\%) \\
				& GraphSage-LR & 0.0843  & 0.1100  & 0.1341  & 0.1567  & 0.1780  \\[-0.4em]
				&              & (21.45\%) & (18.42\%) & (16.26\%) & (14.40\%) & (13.23\%) \\
				& GAT-LR       & 0.0777  & 0.0990  & 0.1194  & 0.1403  & 0.1604  \\[-0.4em]
				&              & (31.76\%) & (31.47\%) & (30.61\%) & (27.78\%) & (25.63\%) \\
				\hline
				F1 Score       & DPA-LR        & 0.0945  & 0.1061  & 0.1149  & 0.1215  & 0.1268  \\
				& GCN-LR       & 0.0760  & 0.0866  & 0.0956  & 0.1029  & 0.1089  \\[-0.4em]
				&              & (24.31\%) & (22.51\%) & (20.24\%) & (18.09\%) & (16.38\%) \\
				& GraphSage-LR & 0.0764  & 0.0873  & 0.0950  & 0.1020  & 0.1068  \\[-0.4em]
				&              & (23.70\%) & (21.62\%) & (20.95\%) & (19.04\%) & (18.71\%) \\
				& GAT-LR       & 0.0726  & 0.0818  & 0.0890  & 0.0952  & 0.1002  \\[-0.4em]
				&              & (30.07\%) & (29.81\%) & (29.16\%) & (27.67\%) & (26.56\%) \\
				
				\hline
			\end{tabular}
			\renewcommand{\arraystretch}{0.8}}
	\end{center}
	\vspace{-3pt}
	\centering{\footnotesize Note: Percentage improvement of our method over a benchmark listed in parentheses.}
\end{table}

\newpage

{\noindent \Large \textbf{Appendix F. Additional Analyses of Our Method }}

\vspace{0.4cm}

{\noindent \large \textbf{F.1 Ranking Quality}}
\vspace{0.4cm}

\noindent We empirically compare the ranking quality between DPA-LR and benchmarks using the metric Discounted Cumulative Gain (DCG), a widely adopted measure for evaluating  the ranking quality of a recommendation list \citep{tan2014object, he2019mobile}. Specifically, the ranking quality of $k$ friends recommended to user $u_i$, measured using $DCG_i$, is given by
\[DCG_i = \sum_{j=1}^{k}\frac{2^{\tau(\mathbb{A}^i, r_j)}-1}{log_2(j+1)} \]
where $r_j$ is the $j^{th}$ ranked friend recommendation in the recommendation list, $\mathbb{A}^i$ denotes the set of friends actually added by $u_i$, and $\tau(\mathbb{A}^i, r_j)$ is an indicator function defined as 
\[\tau(\mathbb{A}^i, r_j) = \begin{cases} 1 & if \  r_j \in \mathbb{A}^i,  \\ 0 & \ otherwise. \end{cases}\]
\noindent According to the definition above, $\tau(\mathbb{A}^i, r_j)$ is 1 if the $j^{th}$ ranked friend recommendation is actually added by $u_i$ and it is 0 otherwise. By the formula of $DCG_i$, the value of $DCG_i$ is larger if a friend recommendation actually added by $u_i$ (i.e., a correct recommendation) ranks higher in the recommendation list. The $DCG$ of a link recommendation method is the average of $DCG_i$ across all users
\[ DCG = \frac{1}{|\mathbb{U}|} \sum_{i=1}^{|\mathbb{U}|}DCG_i, \]
\noindent where $\mathbb{U}$ denotes the set of users. The following table reports the performance comparison between our method (DPA-LR) and benchmarks in terms of $DCG$. As reported in Table A.7, our method substantially and significantly outperforms benchmarks ($p<0.001$). 

\begin{table}[H]
	{\small
		\begin{center}
			\captionsetup{justification=centering,margin=0.5cm}
			\caption{\textbf{Ranking Quality Comparison between DPA-LR and Benchmarks ($k$=10)}}
			\label{tab:table 1}
			\renewcommand{\arraystretch}{0.9}
			\begin{tabular}{c@{\hskip 0.2in}c@{\hskip 0.2in}c@{\hskip 0.2in}} 
				\hline
				\textbf{Method}   & \textbf{DCG} & \textbf{Improvement over Benchmark}     \\
				\hline
				
				DPA-LR                             & 0.7732 &         \\
				MMR($\theta=0.5$)                  & 0.6371 & 21.36\% \\
				MSD($\theta=0.5$)                  & 0.5658 & 36.66\% \\
				DPP($\theta=0.5$)                  & 0.6324 & 22.26\% \\
				DiRec                              & 0.6561 & 17.85\% \\
				
				\hline
			\end{tabular}
			\renewcommand{\arraystretch}{0.9}
			
	\end{center}}
\end{table}


{\noindent \large \textbf{F.2 Sensitivity Analysis}}

\vspace{0.4cm}

\noindent We conduct a sensitivity analysis by varying the number $|\mathbb{C}^i|$ of candidate friends from 50 to 125 with an increment of 25. As reported in Table A.8, our method (i.e., DPA-LR) outperforms each benchmark method substantially and significantly across different numbers of candidate friends ($p<0.001$). As the number of candidate friends increases, more candidates are available for our method to construct a recommendation list that satisfies a user's diversity preference and thus its performance advantage over each benchmark also increases. 
On the other hand, as the number of candidate friends increases, the computational expense of selecting $k$ friends from a larger candidate pool is higher. 
In our main experiments, we set the number of candidate friends to 100 based on the considerations of performance and computational expense.


{\fontfamily{lmss}
	\small{
		\begin{center}
			\textbf{Table A.8 \hspace{0.1cm} Performance Comparison between DPA-LR and Benchmarks ($k$=10): $|\mathbb{C}^i|=50$ to $|\mathbb{C}^i|=125$}
\end{center}}}
\vspace{-4mm}

{\small
\begin{longtable}{c@{\hskip 0.2in} c @{\hskip 0.2in}  c@{\hskip 0.2in} c@{\hskip 0.2in} c @{\hskip 0.2in}c}
		
		\hline 
		\textbf{Metric} &\textbf{Method}&\textbf{$|\mathbb{C}^i|=50$}&\textbf{$|\mathbb{C}^i|=75$}&\textbf{$|\mathbb{C}^i|=100$}&\textbf{$|\mathbb{C}^i|=125$}\\
		\hline 
		\endfirsthead
		
		\multicolumn{6}{c}%
		{{\bfseries \tablename\ \thetable{} {\footnotesize{\fontfamily{lmss}\selectfont -- continued from previous page}}}} \\
		\hline 
		\textbf{Metric} &\textbf{Method}&\textbf{$|\mathbb{C}^i|=50$}&\textbf{$|\mathbb{C}^i|=75$}&\textbf{$|\mathbb{C}^i|=100$}&\textbf{$|\mathbb{C}^i|=125$}\\
		\hline 
		\endhead
		
		\hline 
		\multicolumn{6}{c}{{\footnotesize Continued on next page}} \\
		\endfoot
		\multicolumn{6}{c}{{\footnotesize Note: Percentage improvement of our method over a benchmark listed in parentheses.}} 
		\endlastfoot
	
		DPMS & DPA-LR & 0.3488   & 0.3943   & 0.4559   & 0.4798   \\
		&MMR ($\theta=0.5$)   & 0.2003   & 0.1912   & 0.1813   & 0.1709   \\[-0.4em]
		&	& (74.16\%)  & (106.19\%) & (151.43\%) & (180.75\%) \\
		&MSD ($\theta=0.5$)  & 0.1591   & 0.1383   & 0.1343   & 0.1401   \\[-0.4em]
		&	& (119.21\%) & (185.10\%) & (239.44\%) & (242.47\%) \\
		&DPP ($\theta=0.5$)  & 0.2192   & 0.2092   & 0.1968   & 0.1899   \\[-0.4em]
		&	& (59.10\%)  & (88.48\%)  & (131.64\%) & (152.66\%) \\
		&DiRec & 0.2162   & 0.2129   & 0.2099   & 0.2073   \\[-0.4em]
		&	& (61.30\%)  & (85.21\%)  & (117.27\%) & (131.45\%) \\
		
		\hline 
		Precision & DPA-LR & 0.1517  & 0.1532  & 0.1541 & 0.1537  \\
		& MMR ($\theta=0.5$)   & 0.1291  & 0.1249  & 0.1214 & 0.1191  \\[-0.4em]
		&       & (17.51\%) & (22.66\%) & (26.95\%)  & (29.05\%) \\
		& MSD ($\theta=0.5$)  & 0.1158  & 0.1118  & 0.1114 & 0.1128  \\[-0.4em]
		&       & (31.00\%) & (37.03\%) & (38.30\%)  & (36.26\%) \\
		& DPP ($\theta=0.5$)  & 0.1257  & 0.1221  & 0.1222 & 0.1215  \\[-0.4em]
		&       & (20.68\%) & (25.47\%) & (26.11\%)  & (26.50\%) \\
		& DiRec & 0.1291  & 0.127   & 0.1262 &   0.1258    \\[-0.4em]
		&       & (17.51\%) & (20.63\%) & (22.17\%)  &  (22.17\%)       \\
		\hline
		Recall    & DPA-LR & 0.1463  & 0.1521  & 0.1559 & 0.1573  \\
		& MMR ($\theta=0.5$)  & 0.1139  & 0.1113  & 0.1089  & 0.1078  \\[-0.4em]
		&       & (28.45\%) & (36.66\%) & (43.11\%)  & (45.92\%) \\
		& MSD ($\theta=0.5$)  & 0.1048  & 0.1033  & 0.1035 & 0.1044  \\[-0.4em]
		&       & (39.60\%) & (47.24\%) & (50.56\%)  & (50.67\%) \\
		& DPP ($\theta=0.5$)  & 0.1155  & 0.1153  & 0.1169 & 0.1175  \\[-0.4em]
		&       & (26.67\%) & (31.92\%) & (33.37\%)  & (33.87\%) \\
		& DiRec & 0.1148  & 0.1133  & 0.1128 &   0.1126      \\[-0.4em]
		&       & (27.44\%) & (34.25\%) & (38.19\%)  &   (39.70\%)      \\
		\hline
		F1 Score       & DPA-LR & 0.1109  & 0.1133  & 0.1149 & 0.1152  \\
		& MMR ($\theta=0.5$)  & 0.0917  & 0.0892  & 0.087    & 0.0857  \\[-0.4em]
		&       & (20.94\%) & (27.02\%) & (32.07\%)  & (34.42\%) \\
		& MSD ($\theta=0.5$)  & 0.0829  & 0.081   & 0.0810 & 0.0818  \\[-0.4em]
		&       & (33.78\%) & (39.88\%) & (41.82\%)  & (40.83\%) \\
		& DPP ($\theta=0.5$)  & 0.0892  & 0.0876  & 0.0881 & 0.0881  \\[-0.4em]
		&       & (24.33\%) & (29.34\%) & (30.37\%)  & (30.76\%) \\
		& DiRec & 0.091   & 0.0897  & 0.0892 &    0.0889    \\[-0.4em]
		&       & (21.87\%) & (26.31\%) & (28.89\%)  & (29.58\%) \\
		\hline

\end{longtable}}

\newpage

{\noindent \Large \textbf{Appendix G. Extension to Recommender Systems}}

\vspace{0.4cm}

\noindent We extend our method to recommender systems and evaluate its performance using the MovieLens data set, which contains users' ratings of movies from year 1997 to year 2008 and the features of these movies.\footnote{The data set can be accessed at \url{https://grouplens.org/datasets/hetrec-2011/.}}
Table A.9 gives summary statistics of the data set. As reported, the data set contains 855,598 ratings of 10,197 movies by 2,113 users and a rating ranges from 0.5 (lowest) to 5 (highest). Each movie is described by five features: genre, director, country of origin, filming location, and actor. 
Table A.10 reports summary statistics of movie features. In this table, $n_h$ denotes the number of unique values for feature $h$. For example, there are 20 unique genres in the data set. Notations $max_h$ and $avg_h$ respectively indicate the maximum and average number of values a movie has for feature $h$. For example, a movie in the data set belongs to as many 8 genres and an average of 2.04 genres.

\vspace{-2mm}
\begin{table}[H]
{\small
	\begin{center}
		\captionsetup{justification=centering,margin=0.5cm}
		\caption{\textbf{Summary Statistics of the MovieLens Data Set}}
		\label{tab:table 1}
		\renewcommand{\arraystretch}{0.9}
		\begin{tabular}{ c  @{\hskip 0.15in} c @{\hskip 0.15in} c  @{\hskip 0.15in} c @{\hskip 0.15in} c} 
			\hline
			  \# of Users & \# of Movies & \# of Ratings & \# of Ratings per User & \# of Ratings per Movie \\
			\hline
			  2,113 & 10,197 & 855,598 & 404.92 & 84.64 \\
			\hline
		\end{tabular}
		\renewcommand{\arraystretch}{0.9}
	\end{center}}
\end{table}
\vspace{-1cm}

\begin{table}[H]
{\small
	\begin{center}
		\captionsetup{justification=centering,margin=0.5cm}
		\caption{\textbf{Summary Statistics of Movie Features}}
		\label{tab:table 1}
		\renewcommand{\arraystretch}{0.9}
		\begin{tabular}{c @{\hskip 0.3in} c @{\hskip 0.3in} c @{\hskip 0.3in} c} 
			\hline 
			Feature($h$) & $n_h$ &  $max_h$ & $avg_h$ \\ \hline
			Genre & 20 & 8 & 2.04 \\
			Director & 4,060 & 1 & 1.00 \\
			Country & 72 & 1 & 1.00 \\
			Location & 47,899  & 87 & 4.70 \\
			Actor & 95,321 & 220 & 22.78 \\
			\hline
		\end{tabular}
		\renewcommand{\arraystretch}{0.9}
	\end{center}}
\end{table}
\vspace{-0.6cm}
To apply our DPA-LR method to the MovieLens data set, we need to adapt the definition of diversity preference to the context of movie recommendations. Following the literature \citep{felfernig2018evaluating}, a movie is considered relevant to a user's interest if the user's rating of the movie is above 3.5. A user's diversity preference on feature $h$ (e.g., genre) is measured as the number of the user's relevant movies on each value of the feature. For example, if 8 out of Karen's 10 relevant movies are thrillers, it indicates Karen's strong preference for watching thrillers. On the other hand, if John's relevant movies are evenly distributed over many different genres, it reflects that John prefers a diverse set of genres when watching movies. With diversity preference defined for the context of movie recommendations, we can apply our method to the MovieLens data set. 

Using the MovieLens data set, we evaluated the performance of our method and the benchmark methods listed in Table 4. We divided the data set into training and test data; the training data contained users' movie ratings from year 1997 to year 2006 and the test data contained ratings from year 2007 to year 2008. Similar to the evaluation procedure specified in Section 5.2, we first constructed a candidate set of movie recommendations for each user. Specifically, we employed an accuracy-maximization method -- the Neural Collaborative Filtering (NCF) method \citep{he2017neural}, and trained it with the training data to predict a user's movie ratings in the test period.\footnote{In our evaluation, NCF was implemented using code provided at \url{https://www.kaggle.com/shahrukhkhan/rec-sys-neural-collaborative-filtering-pytorch/notebook}.} A user's candidate set contained top-100 movies with the highest (predicted) ratings by the user, as predicted by the NCF method. Next, each method (ours or benchmark) took a user's candidate set as the input and recommended $k$ movies from the set to the user. The performance of a method was evaluated using the metrics defined in Section 5.2: DPMS, precision, recall, and F1 score. To calculate these metrics, true positive, $TP$, is defined as the number of movies recommended to a user that are actually relevant to the user's interest, where a movie is considered relevant to a user's interest if the user's actual rating of the movie is above 3.5 \citep{felfernig2018evaluating}.
Table A.11 reports the average DPMS, precision, recall, and F1 score of each compared method, averaged across all users. As reported, DPA-LR substantially outperforms each benchmark in all evaluation metrics. The performance advantage of DPA-LR over each benchmark in every evaluation metric is statistically significant ($p<0.001$).

\begin{table}[H]
	{\small
		\begin{center}
			\captionsetup{justification=centering,margin=0cm}
			\caption{\textbf{Performance Comparison between DPA-LR and Benchmarks Using MovieLens Data Set ($k$=10)}}
			\label{tab:table 1}
			\renewcommand{\arraystretch}{0.9}
			\begin{tabular}{c@{\hskip 0.2in}c@{\hskip 0.2in}c@{\hskip 0.2in}c@{\hskip 0.2in}c} 
				\hline
				\textbf{Method} & \textbf{DPMS} & \textbf{Precision} &  \textbf{Recall} &  \textbf{F1 Score}     \\
				\hline  
                DPA-LR & 0.4866  & 0.4321  & 0.0964  & 0.1338  \\
                MMR($\theta=0.5$)    & 0.3954  & 0.3432  & 0.059   & 0.093   \\[-0.4em]
                        & (23.07\%) & (25.90\%) & (63.39\%) & (43.87\%) \\
                MSD($\theta=0.5$)     & 0.4391  & 0.3943  & 0.076   & 0.1164  \\[-0.4em]
                        & (10.82\%) & (9.59\%)  & (26.84\%) & (14.95\%) \\
                DPP($\theta=0.5$)     & 0.4279  & 0.3766  & 0.0744  & 0.1167  \\[-0.4em]
                        & (13.72\%) & (14.74\%) & (29.57\%) & (14.65\%) \\
                DiRec   & 0.4200   & 0.3658  & 0.0663  & 0.1023  \\[-0.4em]
                        & (15.86\%) & (18.12\%) & (45.40\%) & (30.79\%) \\
				\hline
			\end{tabular}
			\renewcommand{\arraystretch}{0.9}
			
	\end{center}}
\vspace{-3pt}
\centering{\footnotesize Note: Percentage improvement of our method over a benchmark listed in parentheses.}
\end{table}
\putbib[DPA-LP]
\end{bibunit}

\end{document}